\pdfoutput=1

\documentclass[11pt]{article}
\usepackage[round]{natbib}
\usepackage{authblk}
\usepackage{times}

\usepackage{graphicx}

\usepackage{amsmath}
\usepackage{array}
\usepackage{algorithmic}
\usepackage[caption=false,subrefformat=parens,labelformat=parens]{subfig}
\usepackage[hyphens]{url}
\usepackage{multirow}
\usepackage{amssymb}
\usepackage{bm}
\usepackage[T1]{fontenc}
\usepackage{appendix}

\usepackage{hyperref}

\DeclareMathOperator{\E}{E}
\DeclareMathOperator{\var}{var}

\DeclareMathOperator{\tr}{tr}
\DeclareMathOperator{\AAssoc}{AA}
\DeclareMathOperator{\RCut}{RC}
\DeclareMathOperator{\NCut}{NC}

\renewcommand{\vec}[1]{\boldsymbol{\mathbf{#1}}}
\newcommand{\bo}{\mathbf}
\newcommand{\cl}{\mathcal}
\newcommand{\dist}{\mathcal D}
\newcommand{\R}{\mathbb R}

\begin{document}

\title{Adaptive Evolutionary Clustering}

\renewcommand\Authfont{\small}
\renewcommand\Affilfont{\small}
\setlength{\affilsep}{3pt}
\author[1]{Kevin S.~Xu}
\author[2]{Mark Kliger}
\author[1]{Alfred O.~Hero III}
\affil[1]{EECS Department, University of Michigan, Ann Arbor, MI, USA \authorcr
\url{xukevin@umich.edu}, \url{hero@umich.edu}}
\affil[2]{Omek Interactive, Israel,
\url{mark.kliger@gmail.com}}

\maketitle

\begin{abstract}
In many practical applications of clustering, the objects to be 
clustered evolve over time, and a clustering result is desired at each 
time step. In such applications, evolutionary 
clustering typically outperforms traditional static clustering 
by producing clustering results that reflect long-term trends 
while being robust to short-term variations. Several 
evolutionary clustering algorithms have recently been proposed, often 
by adding a temporal smoothness penalty to the cost function of a 
static clustering method. In this paper, we introduce a different 
approach to evolutionary clustering by 
accurately tracking the time-varying proximities between 
objects followed by static clustering. We present 
an evolutionary clustering framework that adaptively estimates the optimal 
smoothing parameter using shrinkage estimation, a statistical approach that 
improves a na\"{i}ve estimate using additional information. The proposed 
framework can be used to extend a variety of static clustering algorithms, 
including hierarchical, k-means, and spectral clustering, into evolutionary 
clustering algorithms. 
Experiments on synthetic and real data sets indicate that the 
proposed framework outperforms static clustering and existing 
evolutionary clustering algorithms in many scenarios. 
\end{abstract}

\section{Introduction}
In many practical applications of clustering, the objects to 
be clustered are observed at many points in time, and the goal is to 
obtain a clustering result at each time step. This situation arises in 
applications such as identifying communities in dynamic social 
networks \citep{FalkowskiWI2006,TantipathananandhKDD2007}, tracking groups of 
moving objects \citep{LiKDD2004,Carmi2009}, 
finding time-varying 
clusters of stocks or currencies in financial markets \citep{Fenn2009}, 
and many other applications in data mining, machine learning, 
and signal processing. Typically the objects evolve over time both 
as a result of long-term drifts due to changes in their statistical properties 
and short-term variations due to noise.

A na\"ive approach to these types of problems is to perform static 
clustering at each time step using only 
the most recent data. This approach is extremely 
sensitive to noise and produces clustering results that are unstable 
and inconsistent with clustering results from adjacent time steps. 
Subsequently, evolutionary clustering methods have been developed, with 
the goal of producing clustering results that reflect long-term drifts in the 
objects while being robust to short-term variations\footnote{The term 
``evolutionary clustering'' has also been used to refer to clustering 
algorithms motivated by biological evolution, which are unrelated to the 
methods discussed in this paper.}.

Several evolutionary clustering algorithms have recently been proposed 
by adding a temporal smoothness penalty to the cost function of a static 
clustering method. This penalty prevents the clustering 
result at any given time from deviating too much from the clustering results 
at neighboring 
time steps. This approach has produced evolutionary extensions of commonly 
used static clustering methods such as 
agglomerative hierarchical clustering \citep{ChakrabartiKDD2006}, k-means 
\citep{ChakrabartiKDD2006}, Gaussian mixture models 
\citep{ZhangIJCAI2009}, 
and spectral clustering \citep{TangKDD2008,Chi2009} among 
others. How to choose the weight of the penalty in an optimal manner in 
practice, however, remains an open problem. 

In this paper, we propose a different approach to evolutionary clustering by 
treating it as a problem of tracking followed by static 
clustering (Section \ref{sec:Framework}). 
We model the observed matrix of proximities between objects at each 
time step, which can be either similarities or dissimilarities, as a linear 
combination of a \emph{true proximity matrix} and a zero-mean noise matrix. 
The true proximities, which vary over time, can 
be viewed as \emph{unobserved states} of a dynamic system. Our approach 
involves estimating these states using both current and past proximities, 
then performing static clustering on the state estimates.

The states are estimated using a restricted class 
of estimators known as \emph{shrinkage estimators}, which improve a raw 
estimate by combining it with other information. We develop 
a method for estimating the optimal weight to place on past proximities 
so as to minimize the mean squared error (MSE) between the true 
proximities and our estimates. We call this weight the \emph{forgetting 
factor}. One advantage 
of our approach is that it provides an explicit formula for the optimal 
forgetting factor, unlike existing evolutionary clustering methods. 
The forgetting factor is estimated adaptively, which allows it to vary 
over time to adjust to the conditions of the dynamic system.

The proposed framework, which we call Adaptive Forgetting Factor for 
Evolutionary Clustering and Tracking (AFFECT), can extend any 
static clustering 
algorithm that uses pairwise similarities or dissimilarities into an 
evolutionary clustering algorithm. It is flexible enough 
to handle changes in the number of clusters over time and to accommodate 
objects entering and leaving the data set between time steps. 
We demonstrate how AFFECT can be used to extend 
three popular static clustering algorithms, namely 
hierarchical clustering, k-means, and spectral clustering, into 
evolutionary clustering algorithms (Section \ref{sec:Algorithms}). 
These algorithms are tested on several synthetic and 
real data sets (Section \ref{sec:Experiments}). We find that they not only 
outperform static clustering, but also other recently proposed 
evolutionary clustering algorithms due to the adaptively selected 
forgetting factor.

The main contribution of this paper is the development of the AFFECT 
adaptive evolutionary clustering framework, which has several advantages 
over existing evolutionary clustering approaches:
\begin{enumerate}
	\item It involves smoothing proximities between objects over 
		time followed by static clustering, 
		which enables it to extend any static clustering algorithm 
		that takes a proximity matrix as input to an evolutionary clustering 
		algorithm. 
	\item It provides an explicit formula and estimation 
		procedure for the optimal weight (forgetting factor) to apply to 
		past proximities. 
	\item It outperforms static clustering and existing evolutionary 
		clustering algorithms in several experiments 
		with a minimal increase in computation time 
		compared to static clustering (if a single iteration is used to 
		estimate the forgetting factor).
\end{enumerate}

This paper is an extension of our previous work \citep{XuICASSP2010}, 
which was limited to evolutionary spectral clustering. In this paper, we 
extend the previously proposed framework to other static 
clustering algorithms. 
We also provide additional insight into the model assumptions in 
\citet{XuICASSP2010} and demonstrate the effectiveness of AFFECT in several 
additional experiments.

\section{Background}
\label{sec:Background}

\subsection{Static clustering algorithms}
\label{sec:Static_algs}
We begin by reviewing three commonly used static clustering algorithms. 
We demonstrate the evolutionary extension of these algorithms in Section 
\ref{sec:Algorithms}, although the AFFECT framework 
can be used to extend many other 
static clustering algorithms. 
The term ``clustering'' is used in this paper to refer to both data clustering 
and graph clustering. The notation $i \in c$ is used to denote object $i$ 
being assigned to cluster $c$.
$|c|$ denotes the number of objects in cluster $c$, and $\cl C$ denotes 
a clustering result (the set of all clusters).

In the case of data clustering, we assume that the $n$ objects in the data 
set are stored in an $n \times p$ matrix $X$, where object $i$ is 
represented by a $p$-dimensional feature vector $\vec x_i$ corresponding to 
the $i$th row of $X$. From these feature vectors, one can create a 
proximity matrix $W$, where $w_{ij}$ denotes the 
proximity between objects $i$ and $j$, which could be their Euclidean 
distance or any other similarity or dissimilarity measure.

For graph clustering, we assume that the 
$n$ vertices in the graph are represented by an $n \times n$ adjacency matrix 
$W$ where $w_{ij}$ denotes the weight of the edge between vertices $i$ and 
$j$. If there is no edge between $i$ and $j$, then $w_{ij} = 0$. For the 
usual case of undirected graphs with non-negative edge weights, an adjacency 
matrix is a similarity matrix, so we shall refer to it also as 
a proximity matrix.

\subsubsection{Agglomerative hierarchical clustering}
\label{sec:hierarchical_clustering}
Agglomerative hierarchical clustering algorithms are greedy algorithms that 
create a hierarchical clustering result, often represented by a dendrogram 
\citep{Hastie2001}. The dendrogram can be cut at a certain level to obtain a 
flat clustering result. 
There are many variants of agglomerative hierarchical clustering. A 
general algorithm is described in Fig.~\ref{fig:Hierachical_pseudocode}. 
Varying the definition of dissimilarity between a pair of clusters 
often changes the clustering results. Three common choices are to 
use the minimum dissimilarity between objects in the two clusters 
(single linkage), the maximum dissimilarity (complete 
linkage), or the average dissimilarity (average linkage) \citep{Hastie2001}.

\begin{figure}
	\begin{algorithmic}[1]
		\STATE Assign each object to its own cluster
		\REPEAT
			\STATE Compute dissimilarities between each pair of clusters
			\STATE Merge clusters with the lowest dissimilarity
		\UNTIL{all objects are merged into one cluster}
		\RETURN dendrogram
	\end{algorithmic}
	\caption{A general agglomerative hierarchical clustering algorithm.}
	\label{fig:Hierachical_pseudocode}
\end{figure}

\subsubsection{k-means}
\label{sec:k_means}
k-means clustering \citep{MacQueen1967,Hastie2001} attempts to find 
clusters that minimize the sum of squares cost function
\begin{equation}
	\label{eq:km_cost}
	\dist(X,\cl C) = \sum_{c=1}^k \sum_{i \in c} \|\vec x_i - \vec m_c\|^2,
\end{equation}
where $\|\cdot\|$ denotes the $\ell_2$-norm, and $\vec m_c$ is the centroid 
of cluster $c$, given by
\begin{equation*}
	\vec m_c = \frac{\sum_{i \in c} \vec x_i}{|c|}.
\end{equation*}
Each object is assigned to the cluster with the closest centroid. The cost of 
a clustering result $\cl C$ is simply the sum of squared Euclidean 
distances between 
each object and its closest centroid. 
The squared distance in \eqref{eq:km_cost} can be rewritten as
\begin{equation}
	\label{eq:km_dist_dot_prod}
	\|\vec x_i - \vec m_c\|^2 = w_{ii} - \frac{2\sum_{j \in c} w_{ij}}
		{|c|} + \frac{\sum_{j,l \in c} w_{jl}}{|c|^2},
\end{equation}
where $w_{ij} = \vec x_i \vec x_j^T$, the dot product of the feature 
vectors. Using the form of \eqref{eq:km_dist_dot_prod} to compute the 
k-means cost in \eqref{eq:km_cost} allows the k-means algorithm to be 
implemented with only the similarity matrix $W = [w_{ij}]_{i,j=1}^n$ 
consisting of all pairs of dot products, as described in 
Fig.~\ref{fig:kmeans_pseudocode}. 

\begin{figure}[tp]
	\begin{algorithmic}[1]
		\STATE $i \leftarrow 0$
		\STATE $\cl C^{(0)} \leftarrow$ vector of random integers in 
			$\{1,\ldots,k\}$
		\STATE Compute similarity matrix $W = XX^T$
		\REPEAT
			\STATE $i \leftarrow i+1$
			\STATE Calculate squared distance between all objects and 
				centroids using \eqref{eq:km_dist_dot_prod}
			\STATE Compute $\cl C^{(i)}$ by assigning each object to its 
				closest centroid
		\UNTIL{$\cl C^{(i)} = \cl C^{(i-1)}$}
		\RETURN $\cl C^{(i)}$
	\end{algorithmic}
	\caption{Pseudocode for k-means clustering using similarity matrix $W$.}
	\label{fig:kmeans_pseudocode}
\end{figure}

\subsubsection{Spectral clustering}
\label{sec:spectral_clustering}
Spectral clustering \citep{Shi2000,Ng2001,vonLuxburg2007} 
is a popular modern 
clustering technique inspired by spectral graph theory. It can be used for 
both data and graph clustering. 
When used for data clustering, the first step in spectral clustering is to 
create a similarity graph with vertices corresponding to the objects 
and edge weights corresponding to the similarities between objects. We 
represent the graph by an adjacency matrix $W$ with edge weights $w_{ij}$ 
given by a positive definite similarity function $s(\vec x_i,\vec x_j)$. 
The most commonly used similarity function is the Gaussian 
similarity function $s(\vec x_i,\vec x_j) = \exp\{-\|\vec x_i
- \vec x_j\|^2 / (2\rho^2)\}$ \citep{Ng2001}, where $\rho$ is a 
scaling parameter. 
Let $D$ denote a diagonal matrix with elements corresponding 
to row sums of $W$. Define the unnormalized graph Laplacian matrix by 
$L = D - W$ and the normalized Laplacian matrix \citep{Chung1997} 
by $\cl L = I - D^{-1/2} W D^{-1/2}$.

Three common variants of spectral clustering are average association (AA), 
ratio cut (RC), and normalized cut (NC) \citep{Shi2000}. 
Each variant is associated with 
an NP-hard graph optimization problem. Spectral clustering 
solves relaxed versions of these problems. The relaxed problems 
can be written as \citep{vonLuxburg2007,Chi2009}
\begin{gather}
	\label{eq:AA_spec}
	\AAssoc(Z) = \max_{Z \in \R^{n \times k}} \tr(Z^T W Z) 
		\text{ subject to } Z^T Z = I \\
	\label{eq:RC_spec}
	\RCut(Z) = \min_{Z \in \R^{n \times k}} \tr(Z^T L Z) 
		\text{ subject to } Z^T Z = I \\
	\label{eq:NC_spec}
	\NCut(Z) = \min_{Z \in \R^{n \times k}} \tr(Z^T \cl L Z) 
		\text{ subject to } Z^T Z = I.
\end{gather}
These are variants of a trace optimization problem; the solutions are given 
by a 
generalized Rayleigh-Ritz theorem \citep{Lutkepohl1997}. The optimal solution 
to \eqref{eq:AA_spec} 
consists of the matrix containing the eigenvectors corresponding to the 
$k$ largest eigenvalues of $W$ as columns. Similarly, the optimal 
solutions to \eqref{eq:RC_spec} and \eqref{eq:NC_spec} consist of the 
matrices containing the eigenvectors corresponding to the $k$ smallest 
eigenvalues of $L$ and $\cl L$, respectively. The optimal relaxed 
solution $Z$ is then discretized to obtain a clustering result, 
typically by running the standard k-means algorithm on the rows of $Z$ 
or a normalized version of $Z$.

\begin{figure}[tp]
	\begin{algorithmic}[1]
		\STATE $Z \leftarrow k \text{ smallest eigenvectors of } \cl L$
		\FOR{$i = 1$ to $n$}
		\label{alg:Begin_NC_row_norm}
			\STATE $\bo z_i \leftarrow \bo z_i / \|\bo z_i\|$ 
				\COMMENT{Normalize each row of $Z$ to have unit norm}
		\ENDFOR
		\label{alg:End_NC_row_norm}
		\STATE $\cl C \leftarrow \mathrm{kmeans}(Z)$
		\RETURN $\cl C$
	\end{algorithmic}
	\caption{Pseudocode for normalized cut spectral clustering.}
	\label{fig:Spectral_pseudocode}
\end{figure}

An algorithm \citep{Ng2001} for normalized cut spectral clustering 
is shown in Fig.~\ref{fig:Spectral_pseudocode}.  
To perform ratio cut spectral clustering, compute eigenvectors of $L$ 
instead of 
$\cl L$ and ignore the row normalization in steps 
\ref{alg:Begin_NC_row_norm}--\ref{alg:End_NC_row_norm}. Similarly, 
to perform average association spectral clustering, compute instead the 
$k$ largest eigenvectors of $W$ and ignore the row normalization in steps 
\ref{alg:Begin_NC_row_norm}--\ref{alg:End_NC_row_norm}. 

\subsection{Related work}
We now summarize some contributions in the related areas of incremental 
and constrained clustering, as well as existing work on evolutionary 
clustering. 

\subsubsection{Incremental clustering}
The term ``incremental clustering'' has typically been used to describe 
two types of 
clustering problems\footnote{It is also sometimes used to refer to the 
simple approach of performing static clustering at each time 
step.}:
\begin{enumerate}
	\item Sequentially clustering objects that are each observed only once.
	\label{item:incr_stream}
	\item Clustering objects that are each observed over multiple time 
		steps.
	\label{item:incr_evol}
\end{enumerate}
Type \ref{item:incr_stream} is also known as data stream clustering, and 
the focus is on clustering the data in a single pass and with limited memory 
\citep{Charikar2004,GuptaSDM2004}. It is not directly related to our 
work because in data stream clustering each object is observed only once.

Type \ref{item:incr_evol} is of greater relevance to our work and 
targets the same problem setting as evolutionary clustering. 
Several incremental algorithms of this 
type have been proposed \citep{LiKDD2004,SunKDD2007,NingPattRec2010}. 
These incremental clustering algorithms could 
also be applied to the type of problems we consider; however, the focus 
of incremental clustering is 
on low computational cost at the expense of clustering quality. The 
incremental clustering result is often worse than the result of 
performing static clustering at each time step, which is already a 
suboptimal approach as mentioned in the introduction. On the other 
hand, evolutionary clustering is concerned with improving clustering 
quality by intelligently combining data from multiple time steps and is 
capable of outperforming static clustering.

\subsubsection{Constrained clustering}
The objective of constrained clustering is to find a clustering result that 
optimizes some goodness-of-fit objective (such as the k-means sum of 
squares cost function \eqref{eq:km_cost}) subject to a set of constraints. 
The constraints can either be hard or soft. 
Hard constraints can be used, for example, to specify that two objects must 
or must not be in the same cluster \citep{Wagstaff2001,Wang2010}. 
On the other hand, soft constraints can be used to specify real-valued 
preferences, which may be obtained from labels or other prior information 
\citep{Ji2006,Wang2010}. 
These soft constraints are similar to evolutionary clustering in that they 
bias clustering results based on additional information; in the case of 
evolutionary clustering, the additional information could correspond to 
historical data or clustering results. 

\citet{Tadepalli2009} considered the problem of clustering time-evolving 
objects such that objects in the same cluster at a particular time step 
are unlikely to be in the same cluster at the following time step. 
Such an approach allows one to divide the time series into segments that 
differ significantly from one another. 
Notice that this is the opposite of the evolutionary clustering objective, 
which favors smooth evolutions in cluster memberships over time. 
\citet{Hossain2010} proposed a framework that unifies these two objectives, 
which are referred to as disparate and dependent clustering, respectively. 
Both can be viewed as clustering with soft constraints to minimize or 
maximize similarity between multiple sets of clusters, e.g.~clusters at 
different time steps.

\subsubsection{Evolutionary clustering}
\label{sec:Rel_evol}
The topic of evolutionary clustering has attracted significant attention in 
recent years. \citet{ChakrabartiKDD2006} introduced the 
problem and proposed a general framework for evolutionary clustering 
by adding a temporal smoothness penalty to a static clustering method. 
Evolutionary extensions for agglomerative 
hierarchical clustering and k-means were presented as examples of the 
framework. 

\citet{Chi2009} expanded on this idea by 
proposing two frameworks for evolutionary spectral clustering, which they 
called Preserving Cluster Quality (PCQ) and Preserving Cluster Membership 
(PCM). Both frameworks proposed to optimize the modified cost function 
\begin{equation}
	\label{eq:Evol_clust_cost}
	C_\text{total} = \alpha \, C_\text{temporal} + (1-\alpha) 
		\, C_\text{snapshot},
\end{equation}
where $C_\text{snapshot}$ denotes the static spectral clustering cost, 
which is typically taken to be the average association, ratio cut, 
or normalized cut as discussed in Section 
\ref{sec:spectral_clustering}. The two frameworks differ in how the 
temporal smoothness penalty $C_\text{temporal}$ is defined. In PCQ, 
$C_\text{temporal}$ is defined to be the cost of applying the 
clustering result at time $t$ to the similarity matrix at time $t-1$. In 
other words, it penalizes clustering results that disagree with past 
similarities. In PCM, $C_\text{temporal}$ is defined to be a 
measure 
of distance between the clustering results at time $t$ and $t-1$. In 
other words, it penalizes clustering results that disagree with past 
clustering results. 
Both choices of temporal cost are quadratic in the cluster memberships, 
similar to the static spectral clustering cost as in 
\eqref{eq:AA_spec}--\eqref{eq:NC_spec}, so optimizing 
\eqref{eq:Evol_clust_cost} in either case is simply a trace optimization 
problem. 
For example, the PCQ average association evolutionary spectral clustering 
problem is given by
\begin{equation*}
	\max_{Z \in \R^{n \times k}} \alpha \tr\left(Z^T W^{t-1} 
		Z\right) + (1-\alpha) \tr\left(Z^T W^t Z\right) 
		\text{ subject to } Z^T Z = I \\,
\end{equation*}
where $W^t$ and $W^{t-1}$ denote the adjacency matrices at times $t$ and 
$t-1$, respectively. 
The PCQ cluster memberships can be found by computing eigenvectors of 
$\alpha W^{t-1} + (1-\alpha) W^t$ and then discretizing as 
discussed in Section \ref{sec:spectral_clustering}. 
Our work takes a different approach than that of 
\citet{Chi2009} but the resulting framework shares some 
similarities with the PCQ framework. In particular, AFFECT paired with 
average association spectral clustering is an extension of PCQ to 
longer history, which we discuss in Section \ref{sec:Evol_spectral}.

Following these works, other evolutionary clustering algorithms that 
attempt to optimize the modified cost function defined in 
\eqref{eq:Evol_clust_cost} have been 
proposed \citep{TangKDD2008,LinTKDD2009,ZhangIJCAI2009,MuchaScience2010}. 
The definitions of snapshot and temporal 
cost and the clustering algorithms vary by approach. 
None of the aforementioned works addresses the problem of how to choose 
the parameter $\alpha$ in \eqref{eq:Evol_clust_cost}, which determines how 
much weight to place on historic data or 
clustering results. It has typically been suggested 
\citep{Chi2009,LinTKDD2009} 
to choose it in an ad-hoc manner according to the user's 
subjective preference on 
the temporal smoothness of the clustering results. 

It could also be beneficial to allow $\alpha$ to vary 
with time. \citet{ZhangIJCAI2009} proposed to choose $\alpha$ 
adaptively by using a test statistic for checking dependency between two 
data sets \citep{GrettonAAAI2007}. However, this test statistic also 
does not satisfy any optimality properties for evolutionary clustering 
and still depends on a 
global parameter reflecting the user's preference on temporal smoothness, 
which is undesirable.

The existing method that is most similar to AFFECT is that of 
\citet{RosswogICDMW2008}, which we refer to as RG. 
The authors proposed evolutionary extensions 
of k-means and agglomerative hierarchical clustering by filtering the feature 
vectors using a Finite Impulse 
Response (FIR) filter, which combines the last $l+1$ measurements of 
the feature vectors by the weighted sum 
$\vec y_i^t = b_0 \vec x_i^t + b_1 \vec x_i^{t-1} + \cdots + b_l \vec x_i^{t-l}$,
where $l$ is the order of the filter, $\vec y_i^t$ is the filter output at 
time $t$, and $b_0, \ldots, b_l$ are the filter coefficients. The 
proximities are then calculated between the filter outputs 
rather than the feature vectors. 
The main resemblance between RG and AFFECT is that RG is also based on 
tracking followed by static 
clustering. In particular, RG adaptively selects 
the filter coefficients based on the dissimilarities between cluster 
centroids at the past $l$ time steps. However, RG cannot 
accommodate varying numbers of clusters over time nor can it deal with 
objects entering and leaving at various time steps. 
It also struggles to adapt to changes in clusters, as we demonstrate 
in Section \ref{sec:Coll_Gaussians}. AFFECT, on the other 
hand, is able to adapt quickly to changes in clusters and is applicable to 
a much larger class of problems.

Finally, there has also been recent interest in model-based evolutionary 
clustering. In addition to the aforementioned method involving mixtures of 
exponential families \citep{ZhangIJCAI2009}, methods have also been proposed 
using semi-Markov models \citep{WangSDM2007}, Dirichlet process mixtures 
(DPMs) \citep{AhmedSDM2008,XuICDM2008_2}, hierarchical DPMs 
\citep{XuICDM2008_2,XuICDM2008_1,ZhangKDD2010}, and smooth plaid models 
\citep{Mankad2011}. 
For these methods, the temporal evolution is controlled by 
hyperparameters that can be estimated in some cases. 

\section{Proposed evolutionary framework}
\label{sec:Framework}
The proposed framework treats evolutionary clustering as a tracking problem 
followed by ordinary static clustering. In the case of data clustering, we 
assume that the feature vectors have already been converted into a proximity  
matrix, as discussed in Section \ref{sec:Static_algs}. 
We treat the proximity matrices, denoted by $W^t$, as realizations 
from a non-stationary random process indexed by discrete time steps, denoted 
by the superscript $t$. 
We assume, like many other evolutionary clustering algorithms, that the 
identities of the objects can be tracked over time so that the rows and 
columns of $W^t$ correspond to the same objects as those of $W^{t-1}$ 
provided that no objects are added or removed (we describe how 
the proposed framework handles adding and removing objects in Section 
\ref{sec:Adding_removing}). 
Furthermore we posit the linear observation model
\begin{equation}
	\label{eq:Obs_model}
	W^t = \Psi^t + N^t, \qquad t = 0, 1, 2, \ldots
\end{equation}
where $\Psi^t$ is an unknown deterministic matrix of unobserved states, and 
$N^t$ is a zero-mean noise matrix. $\Psi^t$ changes over time to 
reflect long-term drifts in the proximities. 
We refer to $\Psi^t$ as 
the \emph{true proximity matrix}, and our goal is to accurately estimate it at each 
time step. On the other hand, $N^t$ reflects short-term variations due 
to noise. Thus we assume that $N^t, N^{t-1}, \ldots, N^0$ 
are mutually independent.

A common approach for tracking unobserved states in a dynamic system is to 
use a Kalman filter \citep{Harvey1989,Haykin2001} or some variant. 
Since the states correspond to the true 
proximities, there are $O(n^2)$ states and $O(n^2)$ observations, which makes 
the 
Kalman filter impractical for two reasons. First, it involves specifying a 
parametric model for the 
state evolution over time, and secondly, it requires the inversion of an 
$O(n^2) \times O(n^2)$ covariance matrix, which is large enough in most 
evolutionary clustering applications to make matrix inversion computationally 
infeasible. We present a 
simpler approach that involves a recursive update of the state estimates 
using only a single parameter $\alpha^t$, which we define in 
\eqref{eq:Smoothed_affinity}.

\subsection{Smoothed proximity matrix}
\label{sec:Smoothed_matrix}
If the true proximity matrix $\Psi^t$ is known, we would expect to see 
improved clustering results by performing static clustering on $\Psi^t$ 
rather than on 
the current proximity matrix $W^t$ because $\Psi^t$ is free from noise. 
Our objective is to accurately 
estimate $\Psi^t$ at each time step. We can then perform 
static clustering on our estimate, which should also lead to improved 
clustering results.

The na\"ive approach of performing static clustering on $W^t$ at each time 
step can be interpreted as using $W^t$ itself as 
an estimate for $\Psi^t$. The main disadvantage of this approach is that it 
suffers from high variance due to the observation noise $N^t$. 
As a consequence, the obtained clustering results can be 
highly unstable and inconsistent with clustering results from adjacent 
time steps.

A better estimate can be obtained using the 
\emph{smoothed proximity matrix} $\hat\Psi^t$ defined by 
\begin{equation}
	\label{eq:Smoothed_affinity}
	\hat\Psi^t = \alpha^t \hat\Psi^{t-1} + (1-\alpha^t) W^t
\end{equation}
for $t \geq 1$ and by $\hat\Psi^0 = W^0$. Notice that $\hat\Psi^t$ 
is a function 
of current and past data only, so it can be computed in 
the \emph{on-line} setting where a clustering result for time $t$ is 
desired before data at time $t+1$ can be obtained. 
$\hat\Psi^t$ incorporates proximities not only 
from time $t-1$, but potentially from all previous time steps and allows us 
to suppress the observation noise. The parameter $\alpha^t$ 
controls the rate at which past proximities are forgotten; hence we 
refer to it as the \emph{forgetting factor}. 
The forgetting factor in 
our framework can change over time, allowing the amount of temporal smoothing 
to vary.

\subsection{Shrinkage estimation of true proximity matrix}
\label{sec:Opt_alpha}
The smoothed proximity matrix $\hat\Psi^t$ is a natural candidate for 
estimating $\Psi^t$. 
It is a convex combination of two estimators: $W^t$ and $\hat\Psi^{t-1}$. 
Since $N^t$ is zero-mean, 
$W^t$ is an unbiased estimator but has high variance because it uses only a 
single observation. 
$\hat\Psi^{t-1}$ is a weighted combination of past 
observations so it should have lower variance than $W^t$, but it 
is likely to be biased since the past proximities may not be 
representative of the current ones as a result of long-term drift in the 
statistical properties of the objects. 
Thus the problem of 
estimating the optimal forgetting factor $\alpha^t$ may be considered as a 
bias-variance trade-off problem.

A similar bias-variance 
trade-off has been investigated in the problem of shrinkage estimation of 
covariance matrices \citep{Ledoit2003,Schafer2005,ChenTSP2010}, where 
a shrinkage estimate of the covariance matrix is taken to be $\hat\Sigma 
= \lambda T + (1-\lambda)S$, a convex combination of a suitably chosen target 
matrix $T$ and the standard estimate, the sample covariance matrix $S$. 
Notice that the shrinkage estimate has the same 
form as the smoothed proximity matrix given by 
\eqref{eq:Smoothed_affinity} where the smoothed proximity matrix at 
the previous time step $\hat\Psi^{t-1}$ corresponds to the shrinkage 
target $T$, the current proximity matrix $W^t$ corresponds to 
the sample covariance matrix $S$, and $\alpha^t$ corresponds to the shrinkage 
intensity $\lambda$. 
We derive the optimal choice of $\alpha^t$ in a manner similar to Ledoit and 
Wolf's derivation of the optimal $\lambda$ for shrinkage estimation of 
covariance matrices \citep{Ledoit2003}.

As in \citet{Ledoit2003}, \citet{Schafer2005}, and \citet{ChenTSP2010}, 
we choose to 
minimize the squared Frobenius norm of the difference between the true 
proximity matrix and the smoothed proximity matrix. That is, 
we take the loss function to be
\begin{equation*}
	L\left(\alpha^t\right) = \left\|\hat\Psi^t - \Psi^t\right\|_F^2 
		= \sum_{i=1}^n \sum_{j=1}^n \, \left(\hat\psi_{ij}^t 
		- \psi_{ij}^t\right)^2.
\end{equation*}
We define the risk to be the conditional expectation of the loss function 
given all of the previous observations
\begin{equation*}
	R\left(\alpha^t\right) = \E\left[\left\|\hat\Psi^t - \Psi^t\right\|_F^2 
		\bigg|\,W^{(t-1)}\right]
\end{equation*}
where $W^{(t-1)}$ denotes the set $\left\{W^{t-1}, W^{t-2}, \ldots, 
W^0\right\}$. 
Note that the risk function is differentiable and can be easily 
optimized if $\Psi^t$ is known. However, $\Psi^t$ is the quantity that 
we are trying to estimate so it is not known. We first derive the optimal 
forgetting factor assuming it is known. We shall henceforth refer to this 
as the \emph{oracle forgetting factor}.

Under the linear observation model of \eqref{eq:Obs_model}, 
\begin{gather}
	\label{eq:Cond_mean}
	\E\left[W^t\big|W^{(t-1)}\right] = \E\left[W^t\right] 
		= \Psi^t \\
	\label{eq:Cond_var}
	\var\left(W^t\big|W^{(t-1)}\right) 
		= \var\left(W^t\right) = \var\left(N^t\right)
\end{gather}
because $N^t, N^{t-1}, \ldots, N^0$ are mutually independent and have 
zero mean. From the definition of $\hat\Psi^t$ in 
\eqref{eq:Smoothed_affinity}, 
the risk can then be expressed as
\begin{align}
	R\left(\alpha^t\right) &= \sum_{i=1}^n \sum_{j=1}^n \,
		\E\left[\left(\alpha^t \hat\psi^{t-1}_{ij} + 
		\left(1-\alpha^t\right)w^t_{ij} 
		- \psi^t_{ij}\right)^2 \bigg|\,W^{(t-1)}\right] \nonumber \\
	\begin{split}
		\label{eq:R_var_MS}
		&=\sum_{i=1}^n \sum_{j=1}^n \bigg\{\var\left(\alpha^t 
			\hat\psi^{t-1}_{ij} 
			+ \left(1-\alpha^t\right) w^t_{ij} - \psi^t_{ij}\,\Big|\,W^{(t-1)}
			\right) \\
		&\qquad\qquad\qquad +\E\left[\alpha^t 
			\hat\psi^{t-1}_{ij} + \left(1-\alpha^t\right) w^t_{ij} 
			- \psi^t_{ij}\,\Big|\,W^{(t-1)}\right]^2\bigg\}.
	\end{split}
\end{align}
\eqref{eq:R_var_MS} can be simplified using \eqref{eq:Cond_mean} 
and \eqref{eq:Cond_var} and by noting 
that the conditional variance of $\hat\psi_{ij}^{t-1}$ is zero and 
that $\psi_{ij}^t$ is deterministic. Thus 
\begin{equation}
 	\label{eq:R_simp}
	R\left(\alpha^t\right) = \sum_{i=1}^n \sum_{j=1}^n \left\{\left(
		1-\alpha^t\right)^2 \var\left(n^t_{ij}\right) + \left(\alpha^t
		\right)^2 \left(\hat\psi^{t-1}_{ij} - \psi^t_{ij}\right)^2 \right\}.
\end{equation}
From \eqref{eq:R_simp}, the first derivative is easily seen to be 
\begin{equation*}
	R'\!\left(\alpha^t\right) = 2 \sum_{i=1}^n \sum_{j=1}^n \left\{\left(
		\alpha^t - 1\right)\var\left(n^t_{ij}\right) 
		+ \alpha^t\left(\hat\psi^{t-1}_{ij} - \psi^t_{ij}\right)^2\right\}.
\end{equation*}
To determine the oracle forgetting factor $\left(\alpha^t\right)^*$, simply 
set $R'\!\left(\alpha^t\right) = 0$. Rearranging to isolate $\alpha^t$, we 
obtain
\begin{equation}
	\label{eq:alpha_opt}
	\left(\alpha^t\right)^* = \frac{\displaystyle\sum_{i=1}^n 
		\displaystyle\sum_{j=1}^n \var\left(n^t_{ij}\right)}
		{\displaystyle\sum_{i=1}^n \displaystyle\sum_{j=1}^n 
		\left\{\left(\hat\psi^{t-1}_{ij} - \psi^t_{ij}\right)^2 
		+ \var\left(n^t_{ij}\right)\right\}}.
\end{equation}
We find that 
$\left(\alpha^t\right)^*$ does indeed minimize the risk because 
$R''\!\left(\alpha^t\right) \geq 0$ for all $\alpha^t$.

The oracle forgetting factor $\left(\alpha^t\right)^*$ leads to the best 
estimate in terms of minimizing risk but is not implementable 
because it requires oracle knowledge of 
the true proximity matrix $\Psi^t$, which is what we are trying to 
estimate, as well as the noise variance $\var\left(N^t\right)$. 
It was suggested in \citet{Schafer2005} to replace the unknowns 
with their sample equivalents. In this setting, we would replace 
$\psi_{ij}^t$ with the sample 
mean of $w_{ij}^t$ and $\var(n_{ij}^t) = \var(w_{ij}^t)$ 
with the sample variance of 
$w_{ij}^t$. However, $\Psi^t$ and potentially $\var\left(N^t\right)$ are 
time-varying so we cannot simply use the temporal sample mean and variance.
Instead, we propose to use the \emph{spatial} sample mean and variance. 
Since objects belong to clusters, it is reasonable to assume that the 
structure of $\Psi^t$ and $\var\left(N^t\right)$ should reflect the cluster 
memberships. Hence we make an assumption about the structure of $\Psi^t$ and 
$\var\left(N^t\right)$ in order to proceed.

\subsection{Block model for true proximity matrix}
\label{sec:Block_model}
\begin{figure}[t]
	\centering
	\includegraphics[width=2in]{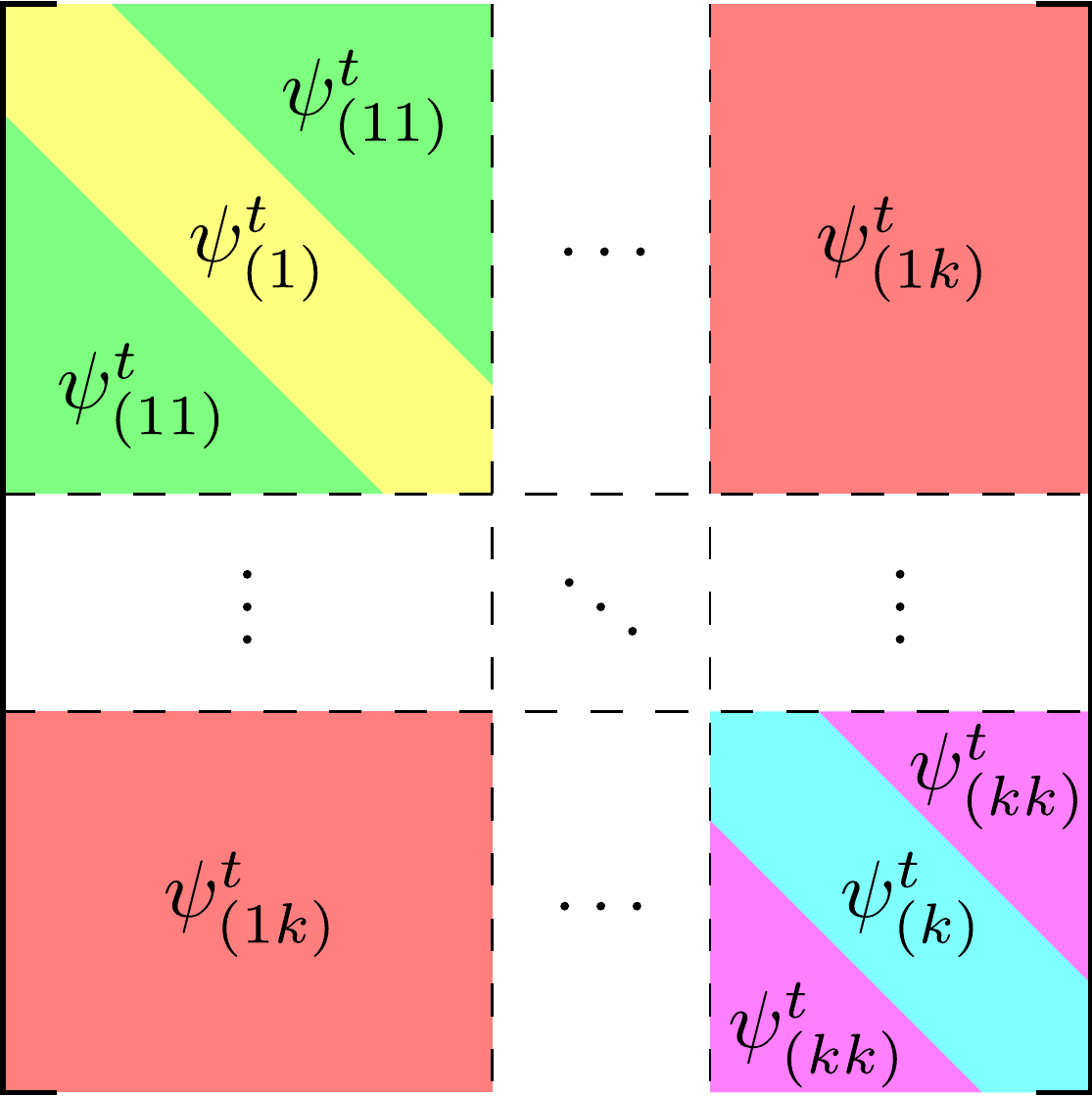}
	\caption{Block structure of true proximity matrix $\Psi^t$. 
		$\psi_{(c)}^t$ denotes $\psi_{ii}^t$ for all objects $i$ in cluster 
		$c$, and $\psi_{(cd)}^t$ denotes $\psi_{ij}^t$ for all 
		distinct objects $i,j$ such that 
		$i$ is in cluster $c$ and $j$ is in cluster $d$.}
	\label{fig:Block_affinity}
\end{figure}

We propose a block model for the true proximity matrix $\Psi^t$ and 
$\var\left(N^t\right)$ and use the assumptions of this model to compute the 
desired sample means and variances.
The assumptions of the block model are as follows:
\begin{enumerate}
	\item $\psi_{ii}^t = \psi_{jj}^t$ for any two objects $i,j$ that 
	belong to the same cluster.
	\item $\psi_{ij}^t = \psi_{lm}^t$ for any two distinct objects $i,j$ 
	and any two distinct objects $l,m$ such that $i,l$ belong to 
	the same cluster, and $j,m$ belong to the same cluster.
\end{enumerate}
The structure of the true proximity matrix $\Psi^t$ under these 
assumptions is shown in Fig.~\ref{fig:Block_affinity}.
In short, we are assuming that the true proximity is equal inside the 
clusters and different between clusters. 
We make the assumptions on $\var\left(N^t\right)$ that we do on $\Psi^t$, 
namely that 
it also possesses the assumed block structure. 

One scenario where the block assumptions are 
completely satisfied is the case where the data at each time $t$ are 
realizations 
from a dynamic Gaussian mixture model (GMM) \citep{Carmi2009}, which is 
described as follows. 
Assume that the $k$ components of 
the dynamic GMM are parameterized by the $k$ time-varying mean vectors 
$\left\{\bm \mu_c^t\right\}_{c=1}^k$ and covariance matrices 
$\left\{\Sigma_c^t\right\}_{c=1}^k$. Let $\left\{\phi_c\right\}_{c=1}^k$ 
denote the mixture weights. Objects are sampled in the following manner:
\begin{enumerate}
	\item (Only at $t=0$) Draw $n$ samples $\left\{z_i\right\}_{i=1}^n$ from 
		the categorical distribution 
		specified by $\left\{\phi_c\right\}_{c=1}^k$ to determine the 
		component membership of each object.
	\item (For all $t$) For each object $i$, draw a sample $\vec x_i^t$ from 
		the Gaussian 
		distribution parameterized by $\left(\bm \mu_{z_i}^t, 
		\Sigma_{z_i}^t\right)$.
\end{enumerate}
Notice that while the parameters of the individual components change over time, 
the component memberships do not, i.e.~objects stay in the same components 
over time. The dynamic GMM simulates clusters moving in time. In 
Appendix \ref{sec:Appendix}, 
we show that at each time $t$, the mean and variance of the dot product 
similarity matrix $W^t$, which correspond to $\Psi^t$ and 
$\var\left(N^t\right)$ respectively under the observation model of 
\eqref{eq:Obs_model}, do indeed satisfy the assumed block structure. This 
scenario forms the basis of the experiment in Section \ref{sec:Sep_Gaussians}.

Although the proposed block model is rather simplistic, we believe that it is 
a reasonable choice when there is no prior information about the shapes of 
clusters. 
A similar block assumption has also been used in the dynamic stochastic 
block model \citep{Yang2011}, developed for modeling dynamic social networks. 
A nice feature of the proposed block model is that it is \emph{permutation 
invariant} with respect to the clusters; that is, it does not require objects 
to be ordered in any particular manner. 
The extension of the proposed framework to other models is 
beyond the scope of this paper and is an area for future work.

\subsection{Adaptive estimation of forgetting factor}
\label{sec:adapt_alpha}
Under the block model assumption, the means and variances of proximities 
are identical in each block. As a result, we can sample over all 
proximities 
in a block to obtain sample means and variances. Unfortunately, we do not 
know the true block structure because the cluster memberships are unknown.

To work around this problem, we estimate the cluster memberships along with 
$\left(\alpha^t\right)^*$ in an iterative fashion. 
First we initialize the cluster memberships. Two logical choices 
are to use the cluster memberships from the previous time step 
or the memberships obtained from performing static clustering on the 
current proximities. We can then sample over each block to estimate the 
entries of $\Psi^t$ and $\var\left(N^t\right)$ as detailed 
below, and substitute them into \eqref{eq:alpha_opt} to obtain an estimate 
$\left(\hat\alpha^t\right)^*$ of $\left(\alpha^t\right)^*$. Now substitute 
$\left(\hat\alpha^t\right)^*$ into \eqref{eq:Smoothed_affinity} and perform 
static clustering on $\hat\Psi^t$ to obtain an updated clustering result. 
This clustering result is then used to refine the estimate of 
$\left(\alpha^t\right)^*$, and this iterative process is repeated to improve 
the quality of the clustering result. 
We find, empirically, that the estimated forgetting factor rarely changes 
after the third iteration and that even a single iteration often provides 
a good estimate.

To estimate the entries of $\Psi^t = \E\left[W^t\right]$, we proceed as 
follows. For two distinct objects $i$ and $j$ both in cluster $c$, we 
estimate $\psi^t_{ij}$ using the sample mean
\begin{equation*}
	\widehat\E\left[w^t_{ij}\right] = \frac{1}{|c|\left(|c|-1\right)} 
		\sum_{l \in c} \sum_{\substack{m \in c\\m \neq l}} w^t_{lm}.
\end{equation*}
Similarly, we estimate $\psi^t_{ii}$ by
\begin{equation*}
	\widehat\E\left[w^t_{ii}\right] = \frac{1}{|c|} \sum_{l \in c} w^t_{ll}.
\end{equation*}
For distinct objects $i$ in cluster $c$ and $j$ in cluster $d$ with 
$c \neq d$, we estimate $\psi^t_{ij}$ by 
\begin{equation*}
	\widehat\E\left[w^t_{ij}\right] = \frac{1}{|c||d|} \sum_{l \in c} 
		\sum_{m \in d} w^t_{lm}.
\end{equation*}
$\var\left(N^t\right) = \var\left(W^t\right)$ can be estimated in a similar 
manner by taking unbiased sample variances over the blocks.

\section{Evolutionary algorithms}
\label{sec:Algorithms}
From the derivation in Section \ref{sec:adapt_alpha}, we have the 
generic algorithm for AFFECT at each time step shown in 
Fig.~\ref{fig:AFFECT_pseudocode}.
We provide some details and interpretation of this generic algorithm when 
used with three popular static clustering algorithms: agglomerative 
hierarchical clustering, k-means, and spectral clustering.

\begin{figure}[tp]
	\begin{algorithmic}[1]
		\STATE $\cl C^t \leftarrow \cl C^{t-1}$
		\FOR[iteration number]{$i=1,2,\ldots$}
			\STATE Compute $\widehat\E\left[W^t\right]$ and 
				$\widehat\var\left(W^t\right)$ using $\cl C^t$
			\STATE Calculate $\left(\hat\alpha^t\right)^*$ by substituting 
				estimates
				$\widehat\E\left[W^t\right]$ and $\widehat\var\left(W^t\right)$ 
				into \eqref{eq:alpha_opt}
			\STATE $\hat \Psi^t \leftarrow \left(\hat\alpha^t\right)^* 
				\hat \Psi^{t-1} + \left[1-\left(\hat\alpha^t\right)^*
				\right] W^t$
			\STATE $\cl C^t \leftarrow \mathrm{cluster}(\hat \Psi^t)$
		\ENDFOR
		\RETURN $\cl C^t$
	\end{algorithmic}
	\caption{Pseudocode for generic AFFECT evolutionary clustering algorithm. 
		$\mathrm{Cluster}(\cdot)$ denotes any static clustering algorithm 
		that takes a similarity or dissimilarity matrix as input and returns 
		a flat clustering result.}
	\label{fig:AFFECT_pseudocode}
\end{figure}

\subsection{Agglomerative hierarchical clustering}
The proposed evolutionary extension of agglomerative hierarchical clustering 
has an interesting interpretation in terms of the modified cost function 
defined in \eqref{eq:Evol_clust_cost}. 
Recall that agglomerative hierarchical clustering 
is a greedy algorithm that merges the two clusters with the lowest 
dissimilarity at each iteration. The dissimilarity between two clusters 
can be interpreted as the cost of merging them. 
Thus, performing 
agglomerative hierarchical clustering on $\hat\Psi^t$ results in 
merging the 
two clusters with the lowest modified cost at each iteration. 
The snapshot cost of a merge corresponds to the cost of making the merge at 
time $t$ using the dissimilarities given by $W^t$. 
The temporal cost of a merge is a weighted combination of the costs of 
making the merge at each time step 
$s \in \{0,1,\ldots,t-1\}$ using the dissimilarities given by $W^s$. 
This can be seen by expanding the recursive update in 
\eqref{eq:Smoothed_affinity} to obtain 
\begin{equation}
	\label{eq:Smoothed_aff_expanded}
	\begin{split}
		\hat\Psi^t = \left(1-\alpha^t\right)W^t &+ \alpha^t \big(1 
			- \alpha^{t-1}\big)W^{t-1} + \alpha^t \alpha^{t-1} 
			\left(1-\alpha^{t-2}\right) W^{t-2} \\
		+ \cdots 
		&+ \alpha^t \alpha^{t-1} \cdots \alpha^2 \left(1-\alpha^1\right)W^1 
			+ \alpha^t \alpha^{t-1} \cdots \alpha^2 \alpha^1 W^0.
	\end{split}
\end{equation}

\subsection{k-means}
k-means is an iterative clustering algorithm and requires an initial 
set of cluster memberships to begin the iteration. In static k-means, 
typically a 
random initialization is employed. A good initialization can significantly 
speed up the algorithm by reducing the number of iterations required for 
convergence. For evolutionary k-means, an obvious choice is to initialize 
using the clustering result at the previous time step. We use this 
initialization in our experiments in Section \ref{sec:Experiments}.

The proposed evolutionary k-means algorithm can also be interpreted as 
optimizing the modified cost function of 
\eqref{eq:Evol_clust_cost}. 
The snapshot cost is $\dist\left(X^t,\cl C^t\right)$ 
where $\dist(\cdot,\cdot)$ is the sum of squares cost defined in 
\eqref{eq:km_cost}. The temporal 
cost is a weighted combination of $\dist\left(X^t,\cl C^s\right), 
s \in \{0,1,\ldots,t-1\}$, i.e.~the cost of the clustering result applied 
to the data at time $s$. Hence the modified cost measures how well the 
current clustering result fits both current and past data.

\subsection{Spectral clustering}
\label{sec:Evol_spectral}
The proposed evolutionary average association spectral clustering 
algorithm involves 
computing and discretizing eigenvectors of $\hat\Psi^t$ rather than $W^t$. It 
can also be interpreted in terms of the modified cost function of 
\eqref{eq:Evol_clust_cost}. Recall that the 
cost in static average association spectral clustering 
is $\tr\left(Z^T W Z\right)$. Performing average association spectral 
clustering on $\hat\Psi^t$ optimizes
\begin{equation}
	\label{eq:Mod_AA_spec_cost}
	\tr\left(Z^T \left[\sum_{s=0}^t \beta^s W^s\right] Z\right) 
		= \sum_{s=0}^t \beta^s \tr\left(Z^T W^s Z\right),
\end{equation}
where $\beta^s$ corresponds to the coefficient in front of $W^s$ in 
\eqref{eq:Smoothed_aff_expanded}. 
Thus, the snapshot cost is simply $\tr\left(Z^T W^t Z\right)$ while the 
temporal cost corresponds to the remaining $t$ terms in 
\eqref{eq:Mod_AA_spec_cost}. 
We note that in the case where $\alpha^{t-1} = 0$, 
this modified cost is identical 
to that of PCQ, which incorporates historical data from time $t-1$ only. 
Hence our proposed generic framework reduces to PCQ in this special 
case.

\citet{Chi2009} noted that PCQ can easily be extended 
to accommodate longer history and suggested to do so by using a constant 
exponentially 
weighted forgetting factor. Our proposed framework uses an adaptive 
forgetting factor, which should improve clustering 
performance, especially if the rate at which the statistical properties 
of the data are evolving is time-varying.

Evolutionary ratio cut and normalized cut spectral clustering can be performed 
by forming 
the appropriate graph Laplacian, $L^t$ or $\cl L^t$, respectively, using 
$\hat\Psi^t$ instead of $W^t$. They do not 
admit any obvious interpretation in terms of a modified cost function since 
they operate on $L^t$ and $\cl L^t$ rather than $W^t$.

\subsection{Practical issues}

\subsubsection{Adding and removing objects over time}
\label{sec:Adding_removing}
Up to this point, we have assumed that the same objects are being observed at 
multiple time steps. In many application scenarios, however, new objects are 
often introduced over time while some existing objects may no longer be 
observed. 
In such a scenario, the indices of the 
proximity matrices $W^t$ and $\hat\Psi^{t-1}$ correspond to different 
objects, so one cannot simply combine them as described in 
\eqref{eq:Smoothed_affinity}.

\begin{figure}
	\centering
	\includegraphics[width=3.3in]{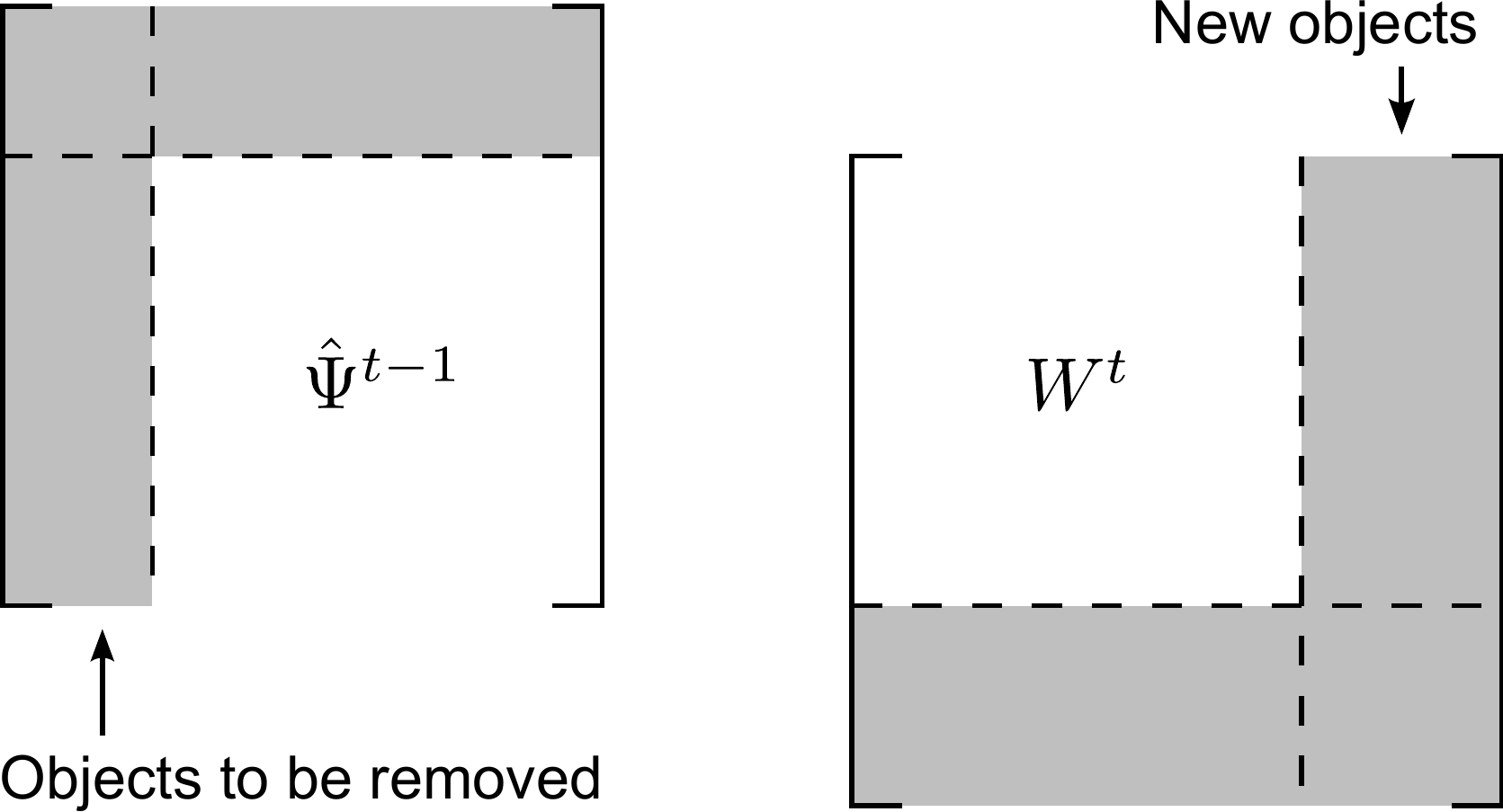}
	\caption{Adding and removing objects over time. 
		Shaded rows and columns are to be removed before computing 
		$\hat\Psi^t$. The rows and columns for the new objects are then 
		appended to $\hat\Psi^t$.}
	\label{fig:Adding_removing}
\end{figure}

These types of scenarios can be dealt with in the following manner. 
Objects that were observed at time $t-1$ but not at time $t$ 
can simply be removed from $\hat\Psi^{t-1}$ in 
\eqref{eq:Smoothed_affinity}. New objects introduced 
at time $t$ have no corresponding rows and columns in $\hat\Psi^{t-1}$.  
These new objects can be naturally handled by adding rows and columns to 
$\hat\Psi^t$ after performing the smoothing operation in 
\eqref{eq:Smoothed_affinity}. In this way, the new nodes have no 
influence on the update of the forgetting factor $\alpha^t$ yet 
contribute to the clustering result through $\hat\Psi^t$.
This process is illustrated graphically in Fig.~\ref{fig:Adding_removing}.

\subsubsection{Selecting the number of clusters}
\label{sec:Num_clusters}
The task of optimally choosing the number of clusters at each time step 
is a difficult model selection problem that is beyond the scope 
of this paper. However, since the proposed framework involves simply 
forming a smoothed proximity matrix followed by static clustering, 
heuristics used for selecting the number of clusters in static clustering 
can also be used with the proposed evolutionary clustering framework. 
One such heuristic applicable to many clustering algorithms is to 
choose the number of clusters to maximize the average silhouette width 
\citep{Rousseeuw1987}. 
For hierarchical clustering, selection of the number of clusters is often 
accomplished using a stopping rule; a review of many such rules can be found 
in \citet{Milligan1985}.
The eigengap heuristic \citep{vonLuxburg2007} and the modularity criterion 
\citep{NewmanPNAS2006} are commonly used heuristics for spectral clustering. 
Any of these 
heuristics can be employed at each time step to choose the number of 
clusters, which can change over time.

\subsubsection{Matching clusters between time steps}
While the AFFECT framework provides a clustering result at each time 
that is consistent with past results, one still faces the challenge of 
matching clusters at time $t$ with those at times $t-1$ and earlier. 
This requires permuting the clusters in the clustering result at time $t$. 
If a one-to-one cluster matching is desired, then the cluster matching 
problem can be formulated as a maximum weight matching between the clusters 
at time $t$ and those at time $t-1$ with weights corresponding to the number 
of common objects between clusters. 
The maximum weight matching can be found in polynomial time using the 
Hungarian algorithm \citep{Kuhn1955}. 
The more general cases of many-to-one (multiple clusters being merged into 
a single cluster) and 
one-to-many (a cluster splitting into multiple clusters) matching are beyond 
the scope of this paper. 
We refer interested readers to \citet{Greene2010} and \citet{Brodka2012}, 
both of which specifically address the cluster matching problem. 

\section{Experiments}
\label{sec:Experiments}
We investigate the performance of the proposed AFFECT framework 
in five experiments involving both synthetic and real data sets. 
Tracking performance is measured in terms of the 
MSE $\E\left[\|\hat\Psi^t - \Psi^t\|_F^2\right]$, which is the 
criterion we seek 
to optimize. Clustering performance is measured by the Rand index 
\citep{Rand1971}, which 
is a quantity between $0$ and $1$ that indicates the 
amount of agreement between a clustering result and a set of labels, which 
are taken to be the ground truth. 
A higher Rand index indicates higher agreement, with a 
Rand index of $1$ corresponding to perfect agreement. 
We run at least one experiment for each of hierarchical clustering, k-means, 
and spectral clustering and compare the performance of AFFECT against three 
recently proposed evolutionary clustering methods discussed in 
Section \ref{sec:Rel_evol}: RG, PCQ, and PCM. 
We run three iterations of AFFECT unless otherwise specified. 

\subsection{Well-separated Gaussians}
\label{sec:Sep_Gaussians}
This experiment is designed to test the tracking ability of AFFECT.
We draw $40$ samples equally from a mixture of two $2$-D Gaussian 
distributions with mean vectors $(4,0)$ and $(-4,0)$ and with both 
covariance matrices equal to $0.1I$. At each time step, the means 
of the two distributions are moved according to a one-dimensional random walk 
in the first dimension with step size $0.1$, and a new sample is drawn with 
the component memberships fixed, as described in Section 
\ref{sec:Block_model}. 
At time $19$, we change the 
covariance matrices to $0.3I$ to test how well the framework can 
respond to a sudden change.

\begin{figure}[tp]
	\centering
 	\includegraphics[width=3.8in]{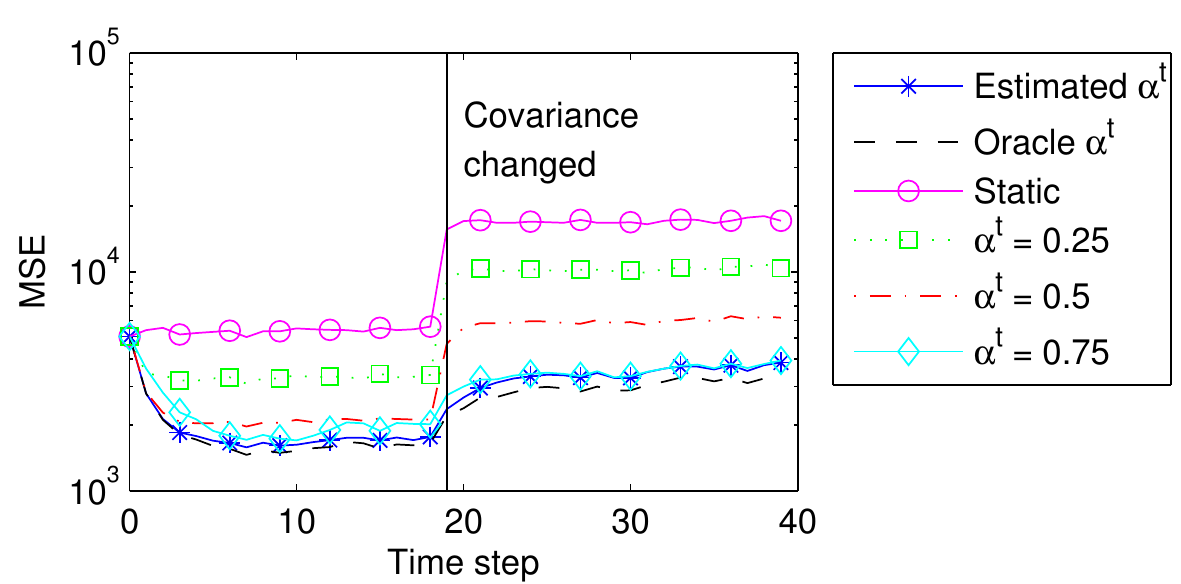}
	\caption{Comparison of MSE in well-separated Gaussians experiment. 
		The adaptively estimated forgetting factor outperforms the constant 
		forgetting factors and achieves MSE very close to the oracle 
		forgetting factor.}
	\label{fig:Sep_Gaussians_MSE}
\end{figure}

We run this experiment $100$ times over $40$ time steps using evolutionary 
k-means clustering. The two clusters 
are well-separated so even static clustering is able to correctly identify 
them. However the tracking performance is improved significantly by 
incorporating historical data, which can be seen in 
Fig.~\ref{fig:Sep_Gaussians_MSE} where the MSE between the estimated and 
true similarity matrices is plotted for several choices of 
forgetting factor, including the estimated $\alpha^t$. 
We also compare to the oracle $\alpha^t$, which 
can be calculated using the true moments and cluster memberships of the data 
as shown in Appendix \ref{sec:Appendix} 
but is not implementable in a real application. Notice that 
the estimated $\alpha^t$ performs very well, and its MSE is very close to that 
of the oracle $\alpha^t$. The estimated $\alpha^t$ also outperforms all 
of the constant forgetting factors.

\begin{figure}[t]
	\centering
	\subfloat[40 samples]{\label{fig:Sep_Gaussians_alpha} 
		\includegraphics[width=1.9in]{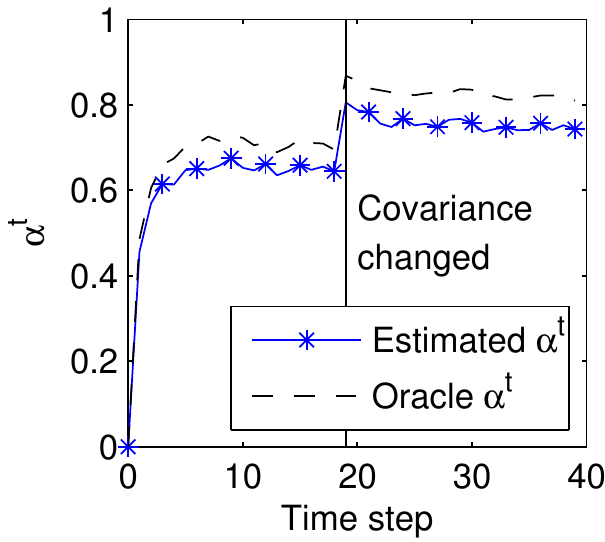}}\qquad\qquad
	\subfloat[200 samples]{\label{fig:Sep_Gaussians_alpha_200} 
		\includegraphics[width=1.9in]{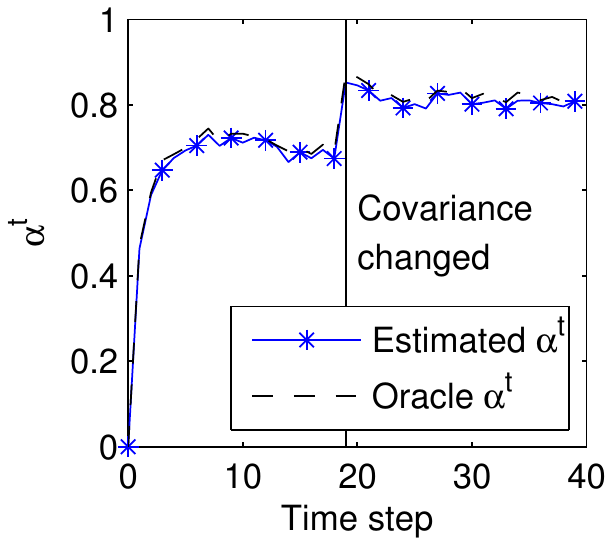}}
	\caption{Comparison of oracle and estimated forgetting factors in 
		well-separated Gaussians experiment. 
		The gap between the estimated 
		and oracle forgetting factors decreases as the sample size 
		increases.}
\end{figure}

The estimated $\alpha^t$ is plotted as a function of time in 
Fig.~\subref*{fig:Sep_Gaussians_alpha}. Since the clusters are well-separated, 
only a single iteration is performed to estimate $\alpha^t$. Notice that 
both the oracle and estimated 
forgetting factors quickly increase from $0$ then level off to a nearly 
constant value until time $19$ when the covariance matrix is changed. After 
the transient due to the change in covariance, both the oracle and estimated 
forgetting factors 
again level off. This behavior is to be expected because the two clusters are 
moving according to random walks. 
Notice that the estimated $\alpha^t$ 
does not converge to the same value the oracle $\alpha^t$ appears to. 
This bias is 
due to the finite sample size. The estimated and oracle forgetting factors 
are plotted in Fig.~\subref*{fig:Sep_Gaussians_alpha_200} for the same 
experiment 
but with $200$ samples rather than $40$. The gap between the steady-state 
values of the estimated and oracle forgetting factors is much smaller now, 
and it continues to decrease as the sample size increases.

\subsection{Two colliding Gaussians}
\label{sec:Coll_Gaussians}
The objective of this experiment is to test the effectiveness of the 
AFFECT framework when a cluster moves close enough to another cluster 
so that they overlap. We also test the ability of the 
framework to adapt to a change in cluster membership.

\begin{figure}[tp]
	\centering
	\includegraphics[width=2.3in]{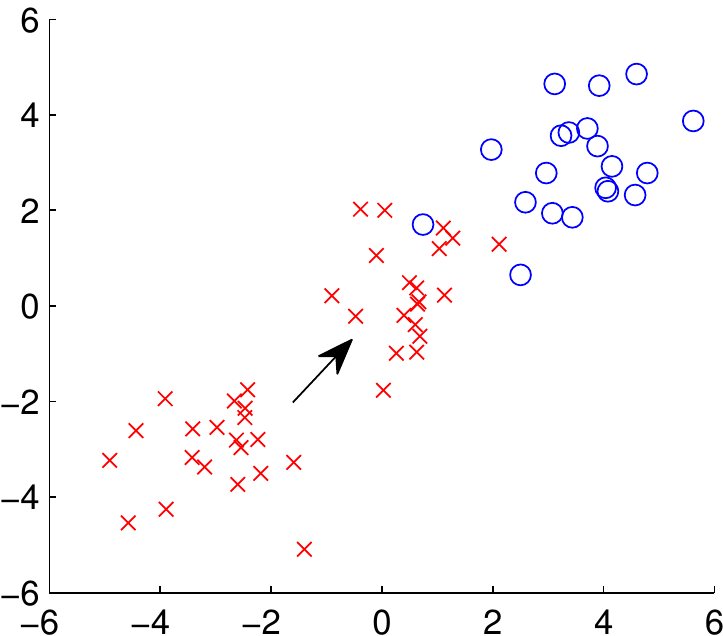}
	\caption{Setup of two colliding Gaussians experiment: one cluster is slowly 
		moved toward the other, then a change in cluster membership 
		is simulated.}
	\label{fig:Two_Gaussians_setup}
\end{figure}

The setup of this experiment is illustrated in Fig. 
\ref{fig:Two_Gaussians_setup}. We draw $40$ samples from a mixture of 
two $2$-D Gaussian distributions, both with covariance matrix equal to 
identity. The mixture proportion (the proportion of samples 
drawn from the second cluster) is initially chosen to be $1/2$. 
The first cluster has mean $(3,3)$ and remains 
stationary throughout the experiment. The second cluster's mean is 
initially at $(-3,-3)$ and is moved toward the first cluster from 
time steps $0$ to $9$ by $(0.4,0.4)$ at each time. At times $10$ and $11$, 
we switch the mixture proportion to $3/8$ and $1/4$, respectively, to 
simulate objects changing cluster. From time $12$ onwards, both the 
cluster means and mixture proportion are unchanged. At each time, we 
draw a new sample.

\begin{figure}[tp]
	\centering
	\includegraphics[width=4.2in]{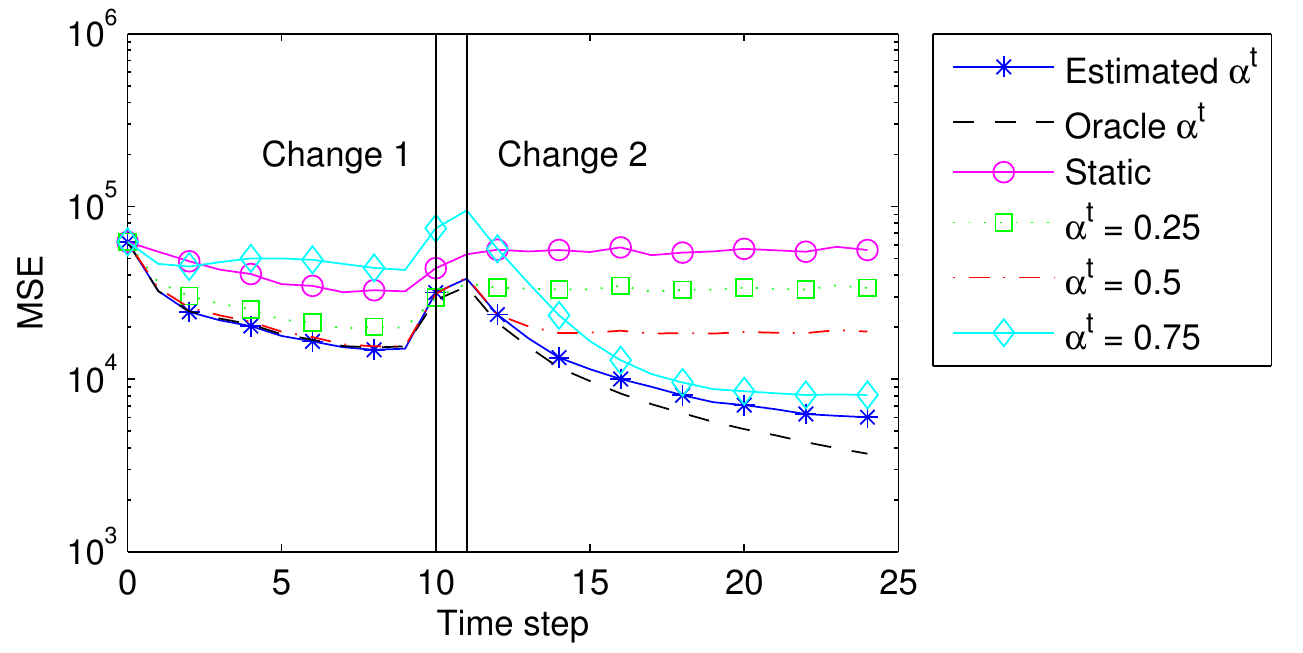}
	\caption{Comparison of MSE in two colliding Gaussians experiment. 
		The estimated 
		$\alpha^t$ performs best both before and after the change points.}
	\label{fig:Two_Gaussians_MSE}
\end{figure}

We run this experiment $100$ times using evolutionary k-means 
clustering. The MSE in this experiment for varying $\alpha^t$ is shown 
in Fig.~\ref{fig:Two_Gaussians_MSE}. As before, the oracle 
$\alpha^t$ is 
calculated using the true moments and cluster memberships and is not 
implementable in practice. It can be seen that the 
choice of $\alpha^t$ affects the MSE significantly. The estimated 
$\alpha^t$ performs the best, excluding the oracle $\alpha^t$, which 
is not implementable. Notice also that $\alpha^t = 0.5$ performs well 
before the change in cluster memberships at time $10$, i.e.~when 
cluster $2$ is moving, while $\alpha^t = 0.75$ performs better after the 
change when both clusters are stationary.

\begin{figure}[tp]
	\centering
	\includegraphics[width=4.2in]{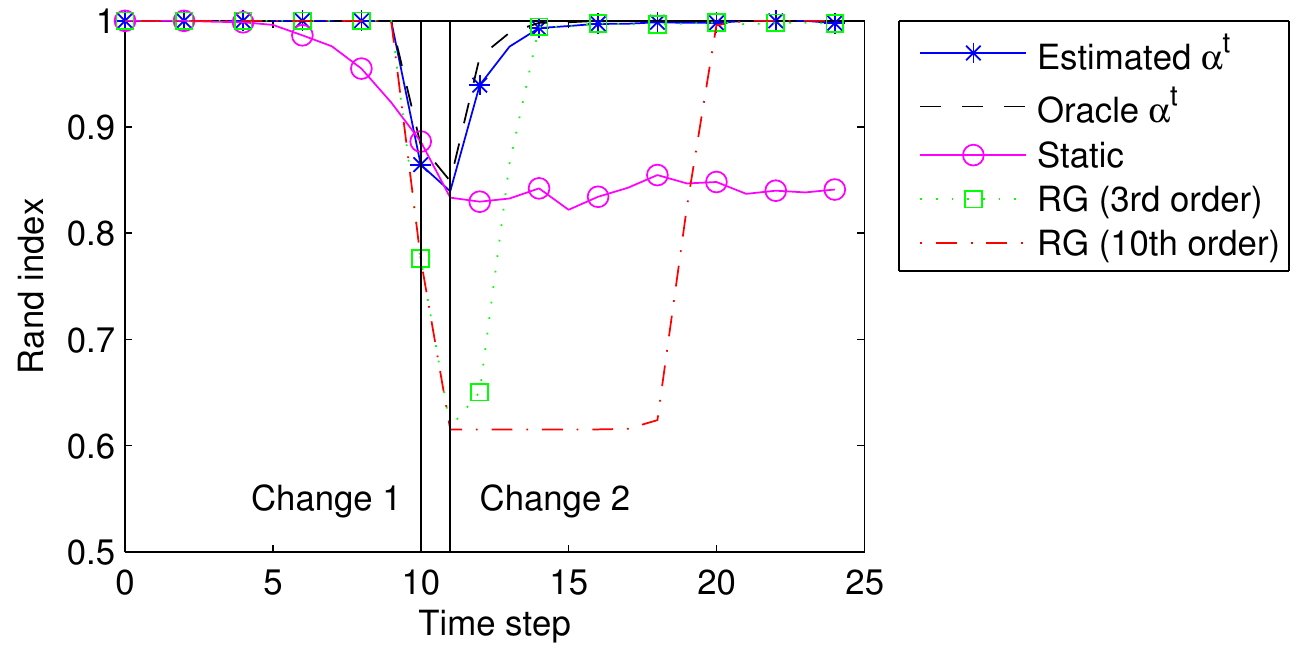}
	\caption{Comparison of Rand index in two colliding Gaussians experiment. 
	The 
	estimated $\alpha^t$ detects the changes in clusters quickly unlike the 
	RG method.}
	\label{fig:Two_Gaussians_Rand}
\end{figure}

The clustering accuracy for this experiment is plotted in 
Fig.~\ref{fig:Two_Gaussians_Rand}. Since this experiment involves k-means 
clustering, we compare to the RG method. 
We simulate two filter lengths for RG: a short-memory $3$rd-order 
filter and a long-memory $10$th-order filter. In 
Fig.~\ref{fig:Two_Gaussians_Rand} it can be seen that the estimated 
$\alpha^t$ also performs best in Rand index, approaching the 
performance of the oracle $\alpha^t$. The static method performs poorly 
as soon as the clusters begin to overlap at around time step $7$. 
All of the evolutionary methods handle the overlap well, but the RG 
method is slow to respond to the change in clusters, especially the 
long-memory filter. In Table \ref{tab:Two_Gaussians_table}, we 
present the means and standard errors (over the simulation runs) of the mean 
Rand indices of each method over all time steps. For AFFECT, 
we also show the Rand index when only one iteration is used to estimate 
$\alpha^t$ and when 
arbitrarily setting $\alpha^t = 0.5$, both of which also 
outperform the RG method in this experiment. The poorer 
performance of the RG method is to be 
expected because it places more weight on time steps where the 
cluster centroids are well-separated, which again results in too much 
weight on historical data after the cluster memberships are changed.

\begin{table}[tp]
	\caption{Means and standard errors of k-means Rand indices in two 
		colliding 
		Gaussians experiment. Bolded number indicates best performer within 
		one standard error.}
	\label{tab:Two_Gaussians_table}
	\setlength{\extrarowheight}{2pt}
	\centering
	\begin{tabular}{ccc}
		\hline
		Method & Parameters & Rand index\\
		\hline
		Static & - & $0.899 \pm 0.002$\\
		\hline
		\multirow{3}{*}{AFFECT} & Estimated $\alpha^t$ ($3$ iterations) 
			& $\bo{0.984 \pm 0.001}$\\
		\cline{2-3}
		& Estimated $\alpha^t$ ($1$ iteration) & $0.978 \pm 0.001$\\
		\cline{2-3}
		& $\alpha^t = 0.5$ & $0.975 \pm 0.001$\\
		\hline
		\multirow{2}{*}{RG} & $l=3$ & $0.955 \pm 0.001$\\
		\cline{2-3}
		& $l=10$ & $0.861 \pm 0.001$\\
		\hline		
	\end{tabular}
\end{table}

\begin{figure}[tp]
	\centering
	\includegraphics[width=4.2in]{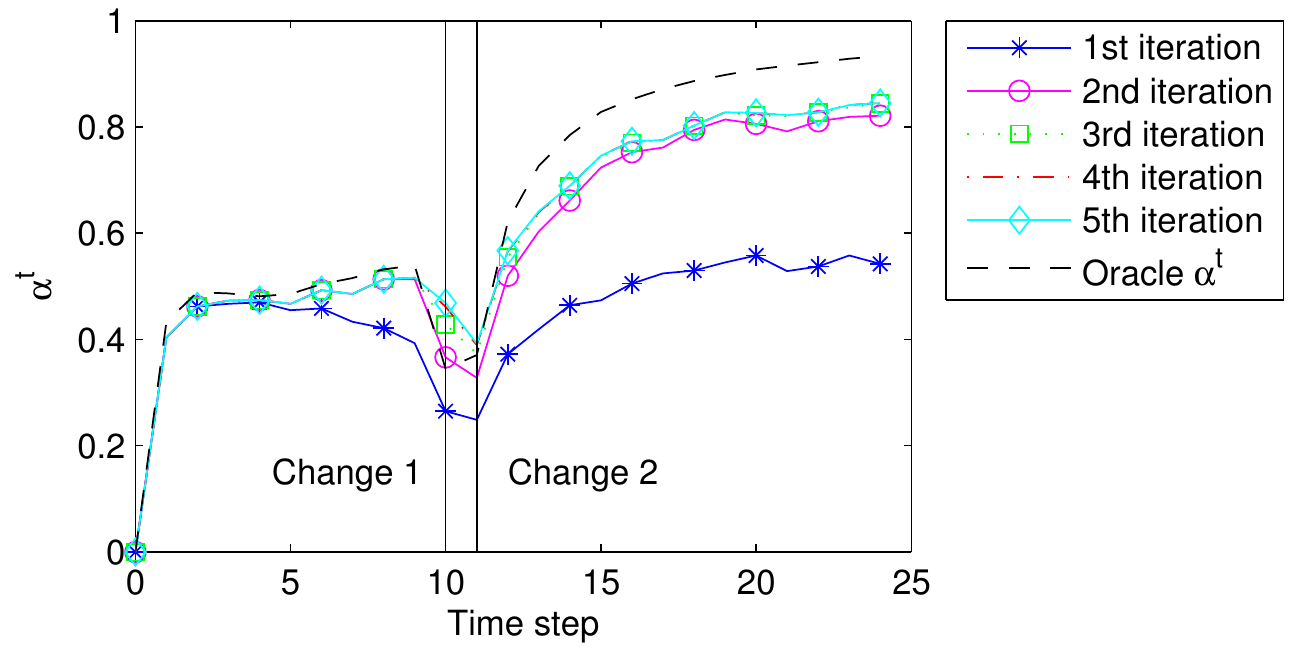}
	\caption{Comparison of oracle and estimated forgetting factors 
		in two colliding Gaussians experiment. 
		There is no noticeable change after the third iteration.}
	\label{fig:Two_Gaussians_alpha}
\end{figure}

The estimated $\alpha^t$ is plotted by 
iteration in Fig.~\ref{fig:Two_Gaussians_alpha} along with the 
oracle $\alpha^t$. 
Notice that the estimate gets better 
over the first three iterations, while the fourth and fifth show no visible 
improvement. 
The plot of the estimated $\alpha^t$ suggests why it is able to 
outperform the constant $\alpha^t$'s. It is almost constant at the beginning 
of the experiment when the second cluster is moving, then it decreases over 
the two times when cluster memberships are changed, and finally it increases 
when the two clusters are both stationary. The values of the oracle $\alpha^t$ 
before and after the change corroborate the previous observation that 
$\alpha^t = 0.5$ performs well before the change, but $\alpha^t = 0.75$ 
performs better afterwards. 
Notice that the 
estimated $\alpha^t$ appears to converge to a lower value than the oracle 
$\alpha^t$. This is once again due to the finite-sample effect discussed in 
Section \ref{sec:Sep_Gaussians}.

\subsection{Flocks of boids}
\label{sec:Boids}
This experiment involves simulation of a natural phenomenon, namely the 
flocking behavior of birds. To simulate this phenomenon we use the bird-oid 
objects (boids) model 
proposed by \citet{ReynoldsSIGGRAPH1987}. The boids model allows us 
to simulate natural movements of objects and clusters. 
The behavior of the boids are governed by three main rules:
\begin{enumerate}
	\item Boids try to fly towards the average position (centroid) of local 
		flock mates.
	\item Boids try to keep a small distance away from other boids.
	\item Boids try to fly towards the average heading of local flock 
		mates.
\end{enumerate}
Our implementation of the boids model is based on the pseudocode of 
\citet{ParkerBoids2007}. At each time step, we move each boid 
$1/100$ of the way towards the average position of local flock mates, 
double the distance between boids that are within $10$ units of each 
other, and move each boid $1/8$ of the way towards the average heading. 

We run two experiments using the boids model; one with a fixed number of 
flocks over time and one where the number of flocks varies over time.

\subsubsection{Fixed number of flocks}

\begin{figure}[tp]
	\centering
	\includegraphics[width=3.7in]{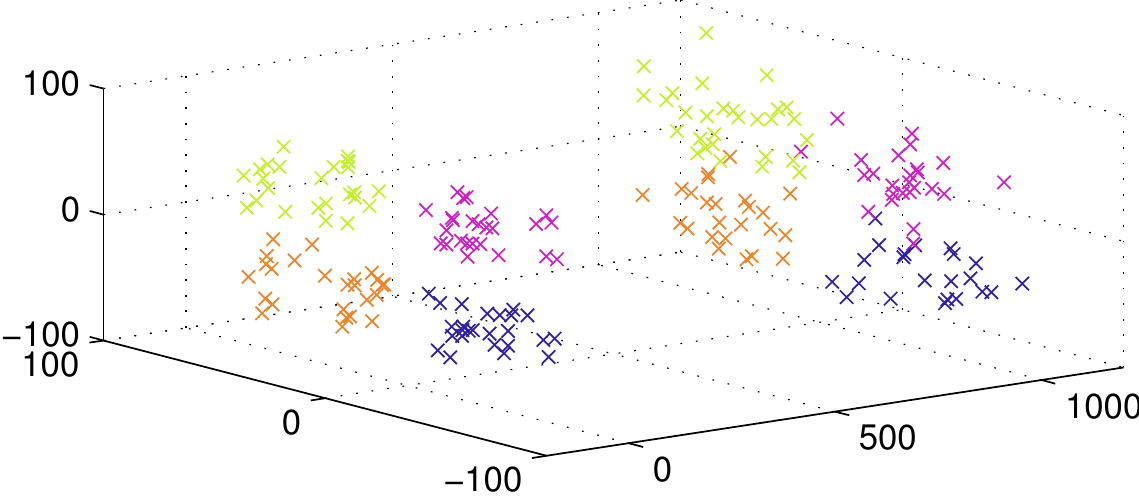}
	\caption{Setup of boids experiment: four flocks fly along parallel 
		paths (start and end positions shown). At each time step, a 
		randomly selected boid joins one of the other flocks.}
	\label{fig:Boids_setup}
\end{figure}

Four flocks of $25$ boids are initially distributed uniformly in 
separate $60 
\times 60 \times 60$ cubes. To simulate boids moving continuously in time 
while being observed at regular time intervals, we allow each boid to 
perform five movements per time step according to the aforementioned 
rules. 
Similar to \citet{ReynoldsSIGGRAPH1987}, we use goal 
setting to push the flocks along parallel paths. 
Note that unlike in the previous experiments, the flocking 
behavior makes it possible to simulate natural changes in cluster, 
simply by changing the flock membership of a boid. 
We change the flock 
memberships of a randomly selected boid at each time step. The initial  
and final positions of the flocks for one realization are shown in 
Fig.~\ref{fig:Boids_setup}.

\begin{figure}[tp]
	\centering
	\includegraphics[width=3.5in]{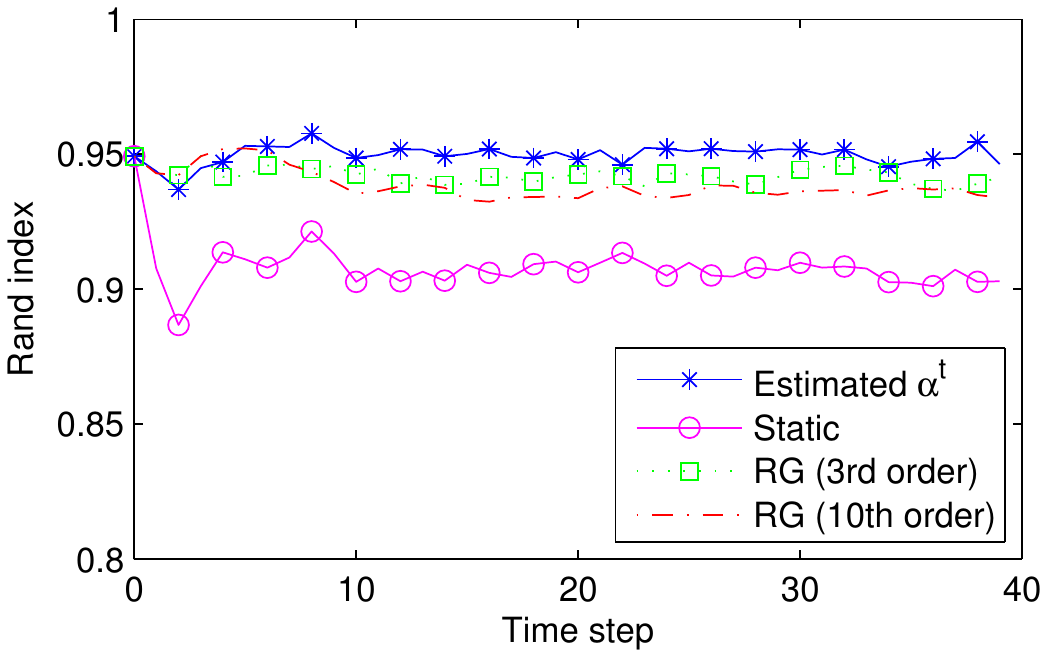}
	\caption{Comparison of complete linkage Rand index in boids experiment. 
		The estimated $\alpha^t$ performs much better than static clustering 
		and slightly better than the RG method.}
	\label{fig:Boids_linkage_Rand}
\end{figure}

\begin{table}[tp]
	\caption{Means and standard errors of complete linkage Rand indices 
		in boids experiment.}
	\label{tab:Boids_linkage_table}
	\setlength{\extrarowheight}{2pt}
	\centering
	\begin{tabular}{ccc}
		\hline
		Method & Parameters & Rand index\\
		\hline
		Static & - & $0.908 \pm 0.001$\\
		\hline
		\multirow{3}{*}{AFFECT} & Estimated $\alpha^t$ ($3$ iterations) 
			& $\bo{0.950 \pm 0.001}$\\
		\cline{2-3}
		& Estimated $\alpha^t$ ($1$ iteration) & $0.945 \pm 0.001$\\
		\cline{2-3}
		& $\alpha^t = 0.5$ & $0.945 \pm 0.001$\\
		\hline
		\multirow{2}{*}{RG} & $l=3$ & $0.942 \pm 0.001$\\
		\cline{2-3}
		& $l=10$ & $0.939 \pm 0.000$\\
		\hline		
	\end{tabular}
\end{table}

We run this experiment $100$ times using complete linkage hierarchical 
clustering. Unlike in the previous experiments, we do not know the 
true proximity matrix so MSE cannot be calculated. Clustering 
accuracy, however, can still be computed using the true flock 
memberships. The clustering performance of the various approaches is 
displayed in Fig.~\ref{fig:Boids_linkage_Rand}. 
Notice that AFFECT once again performs better than RG, both with short and 
long memory, although the difference is much smaller than in the two colliding 
Gaussians experiment. The means and standard errors of 
the Rand indices for the various methods are listed in Table 
\ref{tab:Boids_linkage_table}. Again, it can be seen that AFFECT is 
the best performer. 
The estimated $\alpha^t$ in this experiment is roughly constant at around 
$0.6$. This is not a surprise because 
all movements in this experiment, including changes in clusters, are 
smooth as a result of the flocking motions of the boids. This also explains 
the good performance of simply choosing $\alpha^t = 0.5$ in this 
particular experiment.

\subsubsection{Variable number of flocks}
The difference between this second boids experiment and the first is 
that the number of 
flocks changes over time in this experiment. 
Up to time $16$, this experiment is identical 
to the previous one. At time $17$, we simulate a scattering of the 
flocks by no longer 
moving them toward the average position of local flock mates as well as 
increasing the distance at which boids repel each other to $20$ units. The 
boids are then rearranged at time $19$ into two flocks rather than four. 

We run this experiment $100$ times. 
The RG framework cannot handle changes in the number of clusters over 
time, thus we switch to normalized cut spectral clustering and compare 
AFFECT to PCQ and PCM. 
The number of clusters at each time step is estimated using the modularity 
criterion \citep{NewmanPNAS2006}. 
PCQ and PCM are not equipped with methods for selecting $\alpha$. 
As a result, for each run of the experiment, we first performed a training run 
where the true flock memberships are used to compute the Rand index. The 
$\alpha$ which maximizes the Rand index is then used for the test run. 

\begin{figure}[tp]
	\centering
	\includegraphics[width=3.5in]{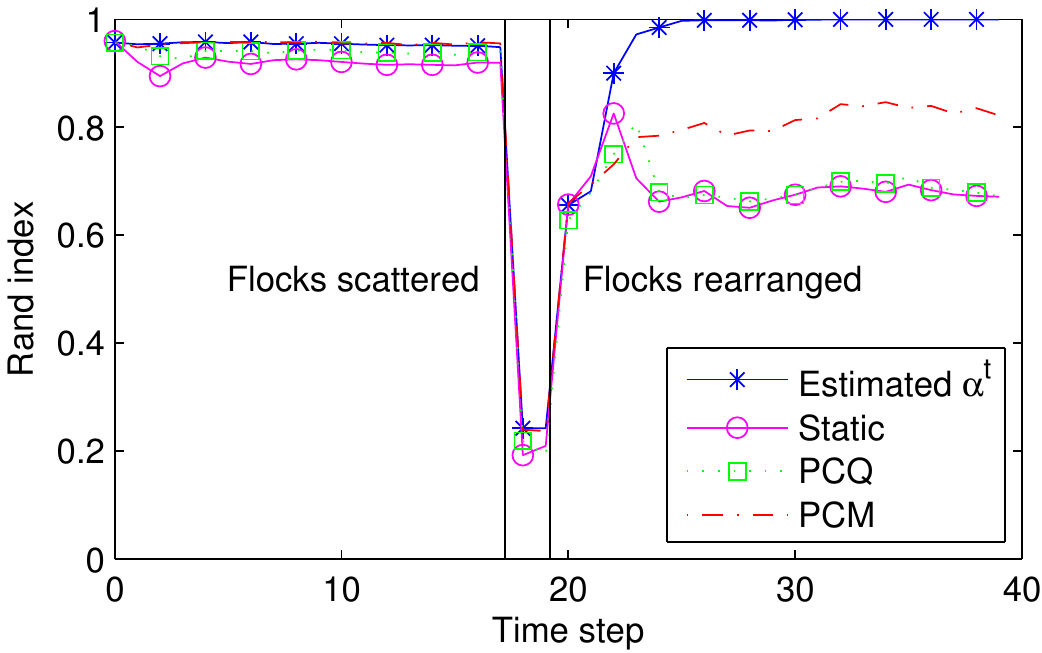}
	\caption{Comparison of spectral clustering Rand index in boids experiment. 
		The estimated $\alpha^t$ outperforms static clustering, PCQ, and 
		PCM.}
	\label{fig:Boids_spectral_Rand}
\end{figure}

\begin{table}[tp]
	\caption{Means and standard errors of spectral clustering Rand indices 
		in boids experiment.}
	\label{tab:Boids_spectral_table}
	\setlength{\extrarowheight}{2pt}
	\centering
	\begin{tabular}{ccc}
		\hline
		Method & Parameters & Rand index\\
		\hline
		Static & - & $0.767 \pm 0.001$\\
		\hline
		\multirow{3}{*}{AFFECT} & Estimated $\alpha^t$ ($3$ iterations) 
			& $\bo{0.921 \pm 0.001}$\\
		\cline{2-3}
		& Estimated $\alpha^t$ ($1$ iteration) & $\bo{0.921 \pm 0.001}$\\
		\cline{2-3}
		& $\alpha^t = 0.5$ & $0.873 \pm 0.002$\\
		\hline
		\multirow{2}{*}{PCQ} & Trained $\alpha$ & $0.779 \pm 0.001$\\
		\cline{2-3}
		& $\alpha = 0.5$ & $0.779 \pm 0.001$\\
		\hline
		\multirow{2}{*}{PCM} & Trained $\alpha$ & $0.840 \pm 0.002$\\
		\cline{2-3}
		& $\alpha = 0.5$ & $0.811 \pm 0.001$\\
		\hline		
	\end{tabular}
\end{table}

\begin{figure}[tp]
	\centering
	\includegraphics[width=3.5in]{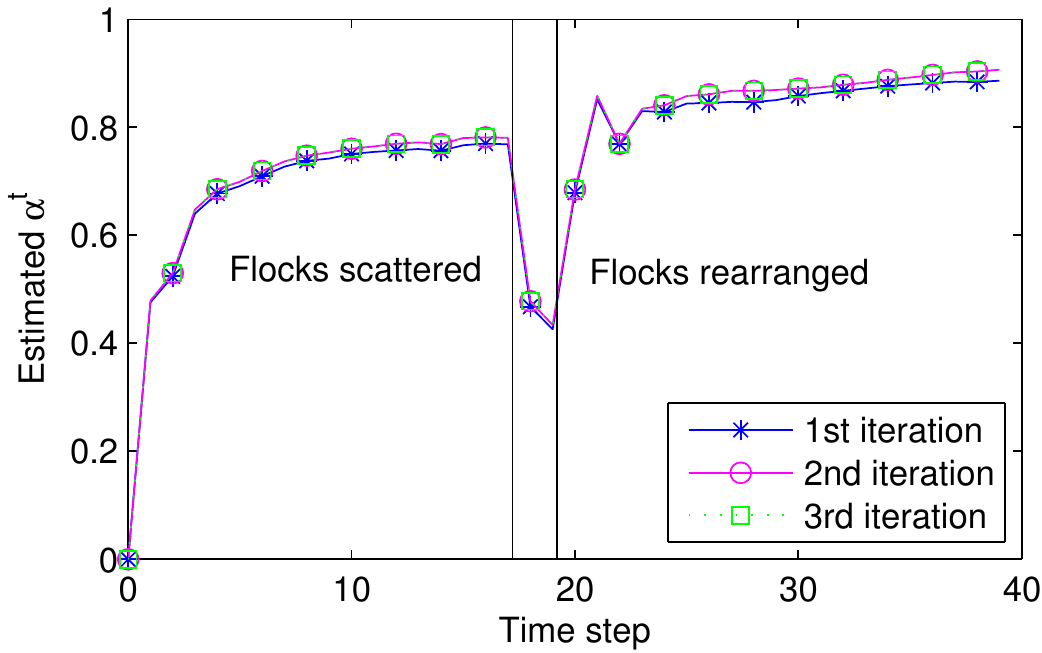}
	\caption{Comparison of estimated spectral clustering forgetting factor 
		by iteration in boids experiment. 
		The estimated forgetting factor 
		drops at the change point, i.e.~when the flocks are scattered. 
		There is no noticeable change in the forgetting factor after the 
		second iteration.}
	\label{fig:Boids_spectral_alpha}
\end{figure}

\begin{figure}[tp]
	\centering
	\includegraphics[width=3.5in]{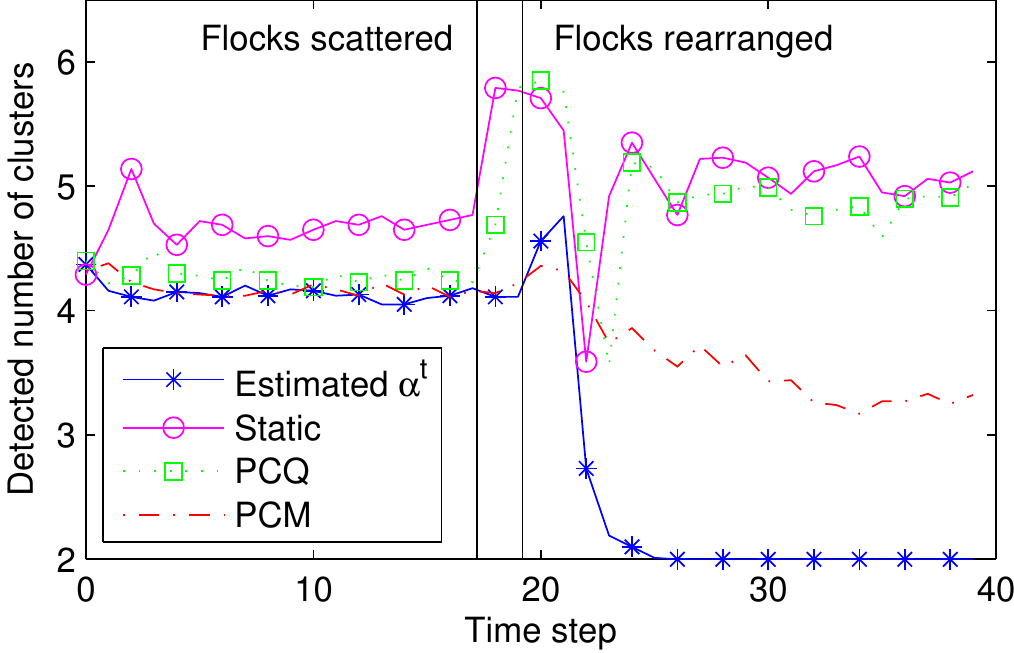}
	\caption{Comparison of number of clusters detected by spectral clustering 
		in boids experiment. 
		Using the estimated $\alpha^t$ results in the 
		best estimates of the number of flocks ($4$ before the change point 
		and $2$ after).}
	\label{fig:Boids_spectral_num_clust}
\end{figure}

The clustering performance is shown in Fig.~\ref{fig:Boids_spectral_Rand}. 
The Rand indices for all methods drop after the flocks are scattered, which is 
to be expected. Shortly after the boids are rearranged into two flocks, the 
Rand indices improve once again as the flocks separate from each other. 
AFFECT once again outperforms the other methods, which can also be seen from 
the summary statistics presented in Table \ref{tab:Boids_spectral_table}. 
The performance of PCQ and PCM with both the trained $\alpha$ and arbitrarily 
chosen $\alpha = 0.5$ are listed. 
Both outperform static clustering but perform noticeably worse than AFFECT 
with estimated $\alpha^t$. 
From Fig.~\ref{fig:Boids_spectral_Rand}, it can be seen that the estimated 
$\alpha^t$ best responds to the rearrangement of the flocks. 
The estimated forgetting factor by iteration is shown in 
Fig.~\ref{fig:Boids_spectral_alpha}.
Notice that the estimated $\alpha^t$ drops when 
the flocks are scattered. 
Notice also that the estimates of $\alpha^t$ 
hardly change after the first iteration, hence why performing one iteration 
of AFFECT achieves the same mean Rand index as performing three iterations. 
Unlike in the previous 
experiments, $\alpha^t = 0.5$ does not perform well in this experiment.

Another interesting observation is that the most accurate 
estimate of the number of clusters at each time is obtained when using 
AFFECT, as shown in 
Fig.~\ref{fig:Boids_spectral_num_clust}. Prior to the flocks being scattered, 
using AFFECT, PCQ, or PCM all result in good estimates for the number of 
clusters, 
while using the static method results in overestimates. However, after the 
rearrangement of the flocks, 
the number of clusters is only accurately estimated when using AFFECT, 
which partially contributes to the poorer Rand indices of PCQ and 
PCM after the rearrangement.

\subsection{MIT Reality Mining}
\label{sec:Reality}
The objective of this experiment is to test the proposed framework 
on a real data set with objects entering and leaving at different 
time steps. The experiment is conducted on 
the MIT Reality Mining data set \citep{EaglePNAS2009}. The data was 
collected by recording cell phone activity of $94$ students and staff 
at MIT over a year. 
Each phone recorded the  
Media Access Control (MAC) addresses of nearby Bluetooth devices at 
five-minute intervals. Using this device proximity data, we construct 
a similarity matrix where the similarity between two students 
corresponds to the number of intervals where they 
were in physical proximity. We divide the data into 
time steps of one week, resulting in $46$ time steps between August 
2004 and June 2005.

In this data set we have partial ground truth. Namely 
we have the affiliations of each participant. \citet{EaglePNAS2009} 
found that two dominant clusters could be 
identified from the Bluetooth proximity data, corresponding to 
new students at the Sloan business school and coworkers who work in the 
same building. 
The affiliations are likely to be 
representative of the cluster structure, at least during the school year. 

\begin{table}[p]
	\caption{Mean spectral clustering Rand indices for MIT Reality Mining
		experiment. Bolded number denotes best performer in each category.}
	\label{tab:Reality_Rand}
	\setlength{\extrarowheight}{2pt}
	\centering
	\begin{tabular}{cccc}
		\hline
		\multirow{2}{*}{Method} & \multirow{2}{*}{Parameters} 
			& \multicolumn{2}{c}{Rand index}\\
		\cline{3-4}
		& & Entire trace & School year \\
		\hline
		Static & - & $0.853$ & $0.905$ \\
		\hline
		\multirow{3}{*}{AFFECT} & Estimated $\alpha^t$ ($3$ iterations) 
			& $\bo{0.893}$ & $\bo{0.953}$ \\
		\cline{2-4}
		& Estimated $\alpha^t$ ($1$ iteration) & $0.891$ & $\bo{0.953}$\\
		\cline{2-4}
		& $\alpha^t = 0.5$ & $0.882$ & $0.949$ \\
		\hline
		\multirow{2}{*}{PCQ} & Cross-validated $\alpha$ & $0.856$ 
			& $0.905$ \\
		\cline{2-4}
		& $\alpha = 0.5$ & $0.788$ & $0.854$ \\
		\hline
		\multirow{2}{*}{PCM} & Cross-validated $\alpha$ & $0.866$ 
			& $0.941$ \\
		\cline{2-4}
		& $\alpha = 0.5$ & $0.554$ & $0.535$ \\
		\hline
	\end{tabular} 
\end{table}

\begin{figure}[p]
	\centering
	\includegraphics[width=4.9in]{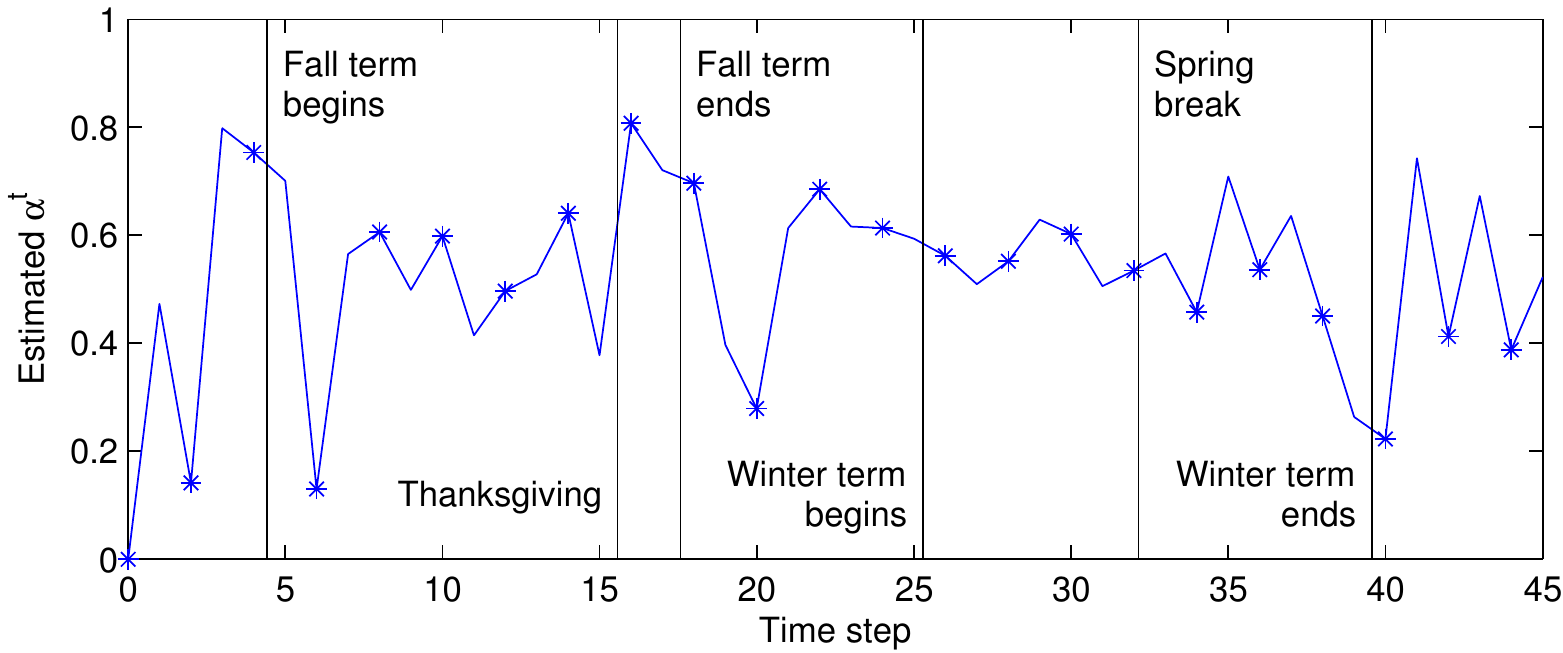}
	\caption{Estimated $\alpha^t$ over entire MIT Reality Mining data trace. 
		Six important dates are indicated. 
		The sudden drops in the 
		estimated $\alpha^t$ indicate change points in the network.}
	\label{fig:Reality_alpha}
\end{figure}

We perform normalized cut spectral clustering into two clusters for this 
experiment and compare AFFECT with PCQ and PCM. 
Since this experiment involves real data, we cannot simulate training sets 
to select $\alpha$ for PCQ and PCM. Instead, we use $2$-fold 
cross-validation, which we believe is the closest substitute. 
A comparison of clustering performance is given in Table 
\ref{tab:Reality_Rand}. 
Both the mean Rand indices over the entire $46$ weeks and only during the 
school year are listed. 
AFFECT is the best performer in both cases. 
Surprisingly, PCQ barely performs better than 
static spectral clustering with the cross-validated $\alpha$ and 
even worse than static spectral clustering with $\alpha = 0.5$. 
PCM fares better than PCQ with the cross-validated $\alpha$ but also performs 
worse than static spectral clustering with $\alpha = 0.5$.
We believe this is 
due to the way PCQ and PCM suboptimally handle objects entering and leaving 
at different 
time steps by estimating previous similarities and memberships, respectively. 
On the contrary, the method used by AFFECT, 
described in Section \ref{sec:Adding_removing}, performs well even with 
objects entering and leaving over time. 

\begin{figure}[tp]
	\centering
	\subfloat{\includegraphics[width=2in]{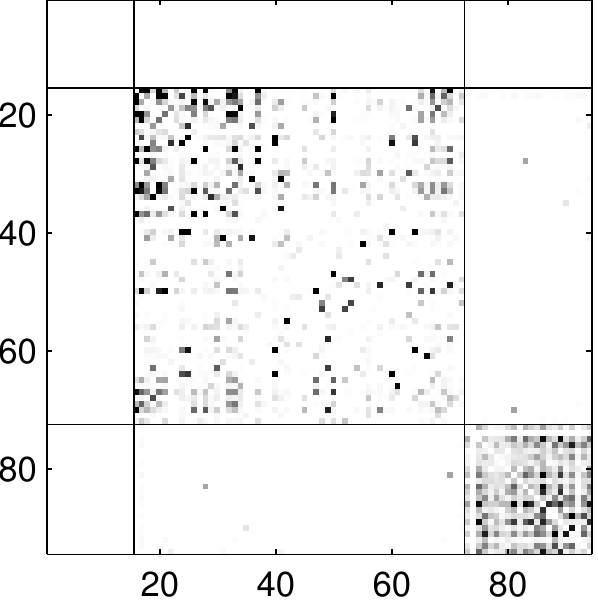}} 
	\qquad\qquad\qquad
	\subfloat{\includegraphics[width=2in]{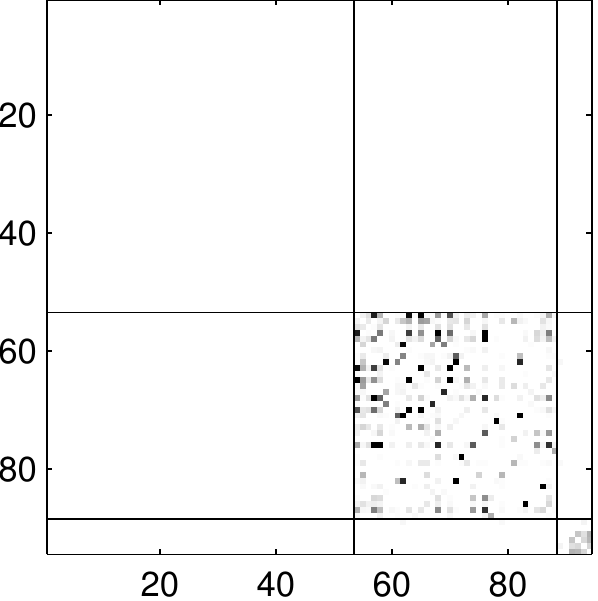}}
	\caption{Cluster structure before (left) and after (right) beginning 
		of winter break in MIT Reality Mining data trace. Darker entries 
		correspond to greater time spent in physical proximity. 
		The empty cluster to the upper left consists of inactive participants 
		during the time step.}
	\label{fig:Reality_clu_break}
\end{figure}

The estimated $\alpha^t$ is shown in Fig.~\ref{fig:Reality_alpha}. 
Six important dates are labeled. The start and end dates of the terms 
were taken from the 
MIT academic calendar \citep{MITCal200405} to be the first and last day 
of classes, respectively.
Notice that the estimated $\alpha^t$ appears to drop around several of 
these dates. These 
drops suggest that physical proximities changed around these dates, 
which is reasonable, especially for the students because their 
physical proximities 
depend on their class schedules. 
For example, the similarity matrices at 
time steps $18$ and $19$, before and after the beginning of winter break,
are shown in Fig.~\ref{fig:Reality_clu_break}. 
The detected clusters using the estimated $\alpha^t$ are superimposed 
onto both matrices, with rows and columns permuted according to the 
clusters. 
Notice that the similarities, corresponding to time spent in physical 
proximity of other participants, are much lower at time $19$, 
particularly in the smaller cluster. 
The change in the structure of the similarity matrix, along with the knowledge 
that the fall term ended and the winter break began around this time, suggests 
that the low estimated forgetting factor at time $19$ is appropriate.

\subsection{NASDAQ stock prices}
In this experiment, we test the proposed framework on a larger time-evolving 
data set, namely stock prices. 
We examined the daily prices of stocks listed on the NASDAQ stock 
exchange in 2008 \citep{Infochimps2012}. 
Using a time step of $3$ weeks ($15$ days in which the stock market is 
operational), we construct a $15$-dimensional vector for each stock 
where the $i$th coordinate consists of the difference between the opening 
prices at the $(i+1)$th and $i$th days. 
Each vector is then normalized by subtracting its sample mean then dividing 
by its sample standard deviation. 
Thus each feature vector $\vec{x}_i^t$ corresponds to the normalized 
derivatives of the opening price sequences over the $t$th $15$-day period. 
This type of feature vector was found by \citet{Gavrilov2000} to provide 
the most accurate static clustering results with respect to the sectors of 
the stocks, which are taken to be the ground truth cluster labels 
\citep{NASDAQ2012}. 
The number of stocks in each sector in the data set for this experiment 
are listed in Table \ref{tab:NASDAQ_sectors}, resulting in a total of 
$2,095$ stocks. 

We perform evolutionary k-means clustering into $12$ clusters, corresponding 
to the number of sectors. 
The mean Rand indices for AFFECT, static clustering, and RG are shown in 
Table \ref{tab:NASDAQ_Rand} along with standard errors over five random 
k-means initializations. 
Since the RG method cannot deal with objects entering and leaving over time, 
we only cluster the $2,049$ stocks listed for the entire year for the 
Rand index comparison. 
AFFECT is once again the best performer, although the improvement is smaller 
compared to the previous experiments. 

\begin{table}[p]
	\caption{Number of stocks in each NASDAQ sector in 2008. 
		The sectors are taken to be the ground truth cluster labels for 
		computing Rand indices.}
	\label{tab:NASDAQ_sectors}
	\setlength{\extrarowheight}{2pt}
	\centering
	\begin{tabular}{c|ccc}
		\hline
		Sector & Basic Industries & Capital Goods & Consumer Durables \\
		\hline
		Stocks & $61$ & $167$ & $188$ \\
		\hline
		\hline
		Sector & Consumer Non-Durables & Consumer Services & Energy \\
		\hline
		Stocks & $93$ & $261$ & $69$ \\
		\hline
		\hline
		Sector & Finance & Health Care & Miscellaneous \\
		\hline
		Stocks & $472$ & $199$ & $65$ \\
		\hline
		\hline
		Sector & Public Utilities & Technology & Transportation \\
		\hline
		Stocks & $69$ & $402$ & $49$ \\
		\hline
	\end{tabular}
\end{table}

\begin{table}[p]
	\caption{Means and standard errors (over five random initializations) of 
		k-means Rand indices for NASDAQ stock prices experiment.}
	\label{tab:NASDAQ_Rand}
	\setlength{\extrarowheight}{2pt}
	\centering
	\begin{tabular}{ccc}
		\hline
		Method & Parameters & Rand index\\
		\hline
		Static & - & $0.801 \pm 0.000$\\
		\hline
		\multirow{3}{*}{AFFECT} & Estimated $\alpha^t$ ($3$ iterations) 
			& $\bo{0.808 \pm 0.000}$\\
		\cline{2-3}
		& Estimated $\alpha^t$ ($1$ iteration) & $0.806 \pm 0.000$\\
		\cline{2-3}
		& $\alpha^t = 0.5$ & $0.806 \pm 0.000$\\
		\hline
		\multirow{2}{*}{RG} & $l=3$ & $0.804 \pm 0.000$\\
		\cline{2-3}
		& $l=10$ & $0.806 \pm 0.001$\\
		\hline		
	\end{tabular}
\end{table}

\begin{figure}[p]
	\centering
	\includegraphics[width=3in]{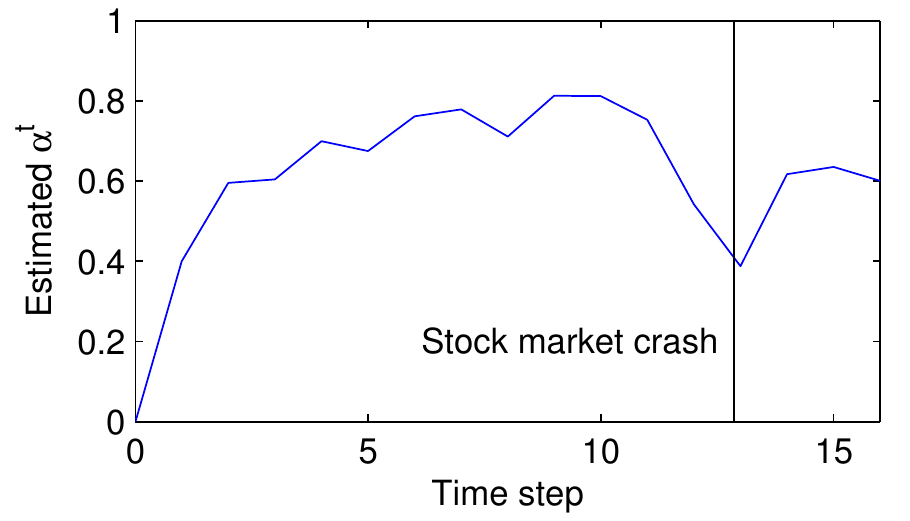}
	\caption{Estimated $\alpha^t$ over NASDAQ stock opening prices in 2008. 
		The sudden drop aligns with the stock market crash in late 
		September.}
	\label{fig:NASDAQ_alpha}
\end{figure}

The main advantage of the AFFECT framework when applied to this data set 
is revealed 
by the estimated $\alpha^t$, shown in Fig.~\ref{fig:NASDAQ_alpha}.
One can see a sudden drop in the estimated $\alpha^t$ at $t=13$ 
akin to the drop seen in the MIT Reality Mining experiment in Section 
\ref{sec:Reality}. 
The sudden drop suggests that there was a significant change in the true 
proximity matrix $\Psi^t$ around this time step, which happens to align 
with the stock market crash that occurred in late September 2008 
\citep{Yahoo2012}, 
once again suggesting the veracity of the downward shift in the value of 
the estimated $\alpha^t$. 

\begin{figure}[tp]
	\centering
	\includegraphics[width=3.5in]{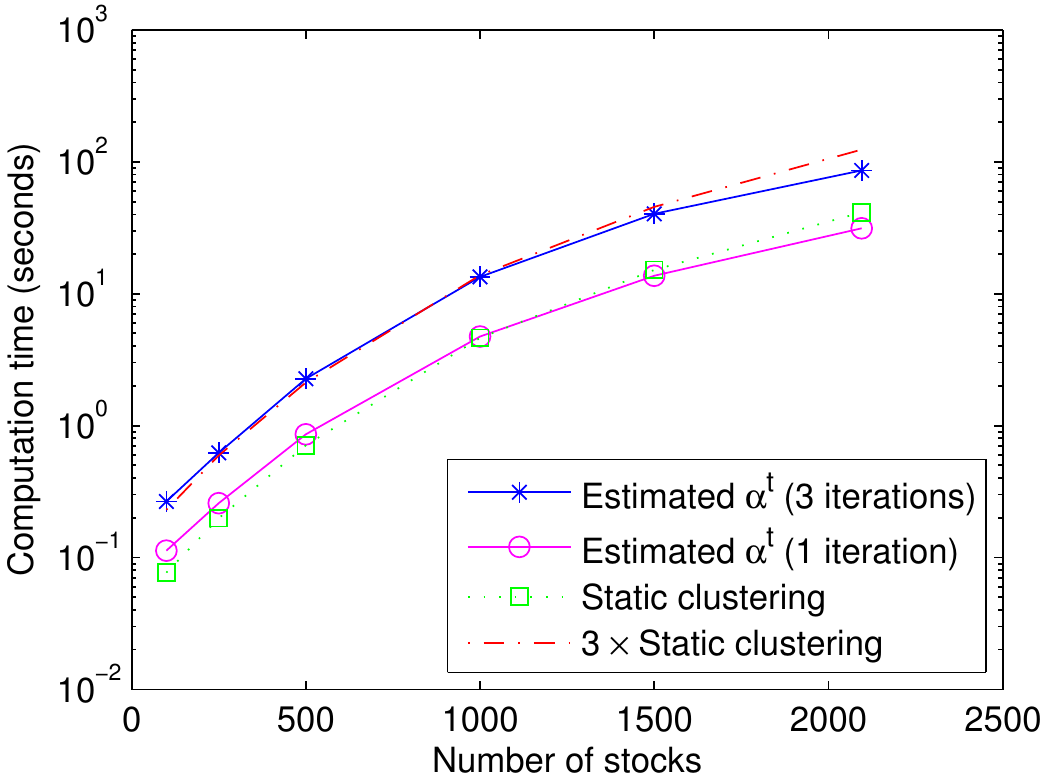}
	\caption{Computation times of AFFECT k-means and static k-means for 
		varying numbers of stocks. 
		The estimation of $\alpha^t$ in AFFECT adds hardly any computation 
		time.}
	\label{fig:NASDAQ_scalability}
\end{figure}

We also evaluate the scalability of the AFFECT framework by varying the 
number of objects to cluster. 
We selected the top $100$, $250$, $500$, $1,000$, and $1,500$ stocks in terms 
of their market cap and compared the computation time of the AFFECT 
evolutionary k-means algorithm to the static k-means algorithm. 
The mean computation times over ten runs on a Linux machine with a 
$3.00\text{ GHz}$ Intel Xeon processor are shown in 
Fig.~\ref{fig:NASDAQ_scalability}. 
Notice that the computation time for AFFECT when running a single iteration 
is almost equivalent to that of static k-means. 
The AFFECT procedure consists of iterating between static clustering and 
estimating $\alpha^t$. 
The latter involves simply computing sample moments over the clusters, which 
adds minimal complexity. 
Thus by performing a single AFFECT iteration, one can achieve better 
clustering performance, as shown in Table \ref{tab:NASDAQ_Rand}, with 
\emph{almost no increase in computation time}.
Notice also that the computation 
time of running a single AFFECT iteration when all $2,095$ stocks are 
clustered is actually \emph{less than} that of static k-means.
This is due to the iterative nature of k-means; clustering on the smoothed 
proximities results in faster convergence of the k-means algorithm. 
As the number of objects increases, the decrease in the computation time due 
to faster k-means convergence is greater than the increase due to estimating 
$\alpha^t$.  
The same observations apply for $3$ iterations of AFFECT when compared to 
$3$ times the computation time for static clustering (labeled as 
``$3 \times \text{static clustering}$''). 

\section{Conclusion}
\label{sec:Conclusion}
In this paper we proposed a novel adaptive framework for evolutionary 
clustering by performing tracking followed by static clustering. The 
objective of the framework was to accurately track the true proximity 
matrix at each time step. This was accomplished using a recursive update 
with an adaptive forgetting factor that controlled the amount of weight to 
apply to historic data. We proposed a method for estimating the optimal 
forgetting factor in order to minimize mean squared tracking error. The main 
advantages of our approach are its universality, allowing almost any static 
clustering algorithm to be extended to an evolutionary one, and that it 
provides an explicit method for selecting the forgetting factor, unlike 
existing methods. 
The proposed framework was evaluated on several synthetic and real data sets 
and displayed good performance in tracking and clustering. It was able 
to outperform both static clustering algorithms and existing evolutionary 
clustering algorithms.

There are many interesting avenues for future work. In the experiments 
presented in this paper, the estimated forgetting factor appeared to converge 
after three iterations. We intend to investigate the convergence properties 
of this iterative process in the future. 
In addition, we would like to improve 
the finite-sample behavior 
of the estimator. Finally, we plan to investigate other loss functions 
and models for the true proximity matrix. We chose to optimize MSE and 
work with a block model in this paper, but perhaps other functions or models 
may be more appropriate for certain applications.

\appendixtitleon
\appendixtitletocon
\begin{appendices}
\section{True similarity matrix for dynamic Gaussian mixture model}
\label{sec:Appendix}
We derive the true similarity matrix $\Psi$ and the matrix of variances 
of similarities $\var(W)$, where the similarity is taken to be the dot 
product, for data sampled from the dynamic Gaussian mixture model 
described in Section \ref{sec:Block_model}. These matrices 
are required in order to calculate the oracle forgetting factor for the 
experiments in Sections \ref{sec:Sep_Gaussians} and 
\ref{sec:Coll_Gaussians}. We drop the superscript $t$ to simplify the 
notation.

Consider two arbitrary objects 
$\vec x_i \sim N(\bm \mu_c,\Sigma_c)$ and $\vec x_j \sim 
N(\bm \mu_d, \Sigma_d)$ where the entries of $\bm \mu_c$ 
and $\Sigma_c$ are denoted by $\mu_{ck}$ and $\sigma_{ckl}$, respectively. 
For any distinct $i,j$ the mean is
\begin{equation*}
	\E\left[\vec x_i \vec x_j^T\right] = \sum_{k=1}^p 
		\E\left[x_{ik}x_{jk}\right] = \sum_{k=1}^p \mu_{ck} \mu_{dk},
\end{equation*}
and the variance is
\begin{align*}
	\var\left(\vec x_i \vec x_j^T\right) &= \E\left[\left(\vec x_i 
		\vec x_j^T\right)^2\right] - \E\left[\vec x_i 
		\vec x_j^T\right]^2 \\
	&= \sum_{k=1}^p \sum_{l=1}^p \left\{\E\left[x_{ik}x_{jk}x_{il}x_{jl}
		\right] - \mu_{ck}\mu_{dk}\mu_{cl}\mu_{dl}\right\} \\
	&= \sum_{k=1}^p \sum_{l=1}^p \left\{\left(\sigma_{ckl} 
		+ \mu_{ck}\mu_{cl}\right) 
		\left(\sigma_{dkl} + \mu_{dk}\mu_{dl}\right) 
		- \mu_{ck}\mu_{dk}\mu_{cl}\mu_{dl}\right\} \\
	&= \sum_{k=1}^p \sum_{l=1}^p \left\{\sigma_{ckl}\sigma_{dkl} 
		+ \sigma_{ckl}\mu_{dk}\mu_{dl} + \sigma_{dkl}\mu_{ck}
		\mu_{cl}\right\}
\end{align*}
by independence of $\vec x_i$ and $\vec x_j$. This holds both for 
$\vec x_i,\vec x_j$ in the same cluster, i.e.~$c=d$, and for $\vec x_i, 
\vec x_j$ in different clusters, i.e.~$c \neq d$. Along the diagonal, 
\begin{equation*}
	\E\left[\vec x_i \vec x_i^T\right] = \sum_{k=1}^p 
		\E\left[x_{ik}^2\right] = \sum_{k=1}^p \left(\sigma_{ckk} 
		+ \mu_{ck}^2\right).
\end{equation*}
The calculation for the variance is more involved. We first note that 
\begin{equation*}
	\E\left[x_{ik}^2 x_{il}^2\right] = \mu_{ck}^2\mu_{cl}^2 
		+ \mu_{ck}^2\sigma_{cll} + 4\mu_{ck}\mu_{cl}\sigma_{ckl}
		+ \mu_{cl}^2\sigma_{ckk} + 2\sigma_{ckl}^2 
		+ \sigma_{ckk}\sigma_{cll},
\end{equation*}
which can be derived from the characteristic function of the multivariate 
Gaussian distribution \citep{Anderson2003}. Thus
\begin{align*}
	\var\left(\vec x_i \vec x_i^T\right) &= \sum_{k=1}^p \sum_{l=1}^p 
		\left\{\E\left[x_{ik}^2 x_{il}^2\right] - \left(\sigma_{ckk} 
		+ \mu_{ck}^2\right) 
		\left(\sigma_{cll} + \mu_{cl}^2\right)\right\} \\
	& = \sum_{k=1}^p \sum_{l=1}^p \left\{4\mu_{ck}\mu_{cl}\sigma_{ckl} 
		+ 2\sigma_{ckl}^2\right\}.
\end{align*}
The calculated means and variances are then substituted into 
\eqref{eq:alpha_opt} to compute the oracle forgetting factor. Since the 
expressions for the means and variances depend only on the clusters and 
not any objects in particular, it is confirmed that 
both $\Psi$ and $\var(W)$ do indeed possess the assumed block structure 
discussed in Section \ref{sec:Block_model}.
\end{appendices}

\section*{Acknowledgements}
We would like to thank the anonymous reviewers for their suggestions to 
improve this article.
This work was partially supported by the 
National Science Foundation grant CCF 0830490 and the US Army Research Office 
grant number W911NF-09-1-0310. 
Kevin Xu was partially supported by an award from the Natural 
Sciences and Engineering Research Council of Canada.

\bibliographystyle{abbrvnat}
\bibliography{full,library_s,references}

\end{document}